\pdfoutput=1
\documentclass[11pt]{article}

\usepackage{authblk}
\usepackage{acl}

\usepackage{times}
\usepackage{latexsym}

\usepackage[T1]{fontenc}
\usepackage[utf8]{inputenc}
\usepackage{enumitem}

\usepackage{microtype}

\usepackage{inconsolata}

\usepackage{graphicx}
\usepackage{caption}
\usepackage{subcaption}
\usepackage{tikz} % for dashed lines
\usepackage{authblk}

\title{Harnessing the Power of Artificial Intelligence to Vitalize \\ Endangered Indigenous Languages: Technologies and Experiences}

\author[1,2]{Claudio Pinhanez}
\author[1]{Paulo Cavalin}
\author[2]{Luciana Storto}
\author[2]{Thomas Finbow} 
\author[2]{\authorcr Alexander Cobbinah}
\author[1]{Julio Nogima}
\author[1]{Marisa Vasconcelos}
\author[1]{\authorcr Pedro Domingues}
\author[1]{Priscila de Souza Mizukami}
\author[2]{Nicole Grell}
\author[2]{\authorcr Majoí Gongora}
\author[1]{Isabel Gonçalves}

\affil[1]{IBM Research Brazil}
\affil[2]{University of S\~ao Paulo}
\begin{document}
\maketitle

\begin{abstract}
Since 2022 we have been exploring application areas and technologies in which Artificial Intelligence (AI) and modern Natural Language Processing (NLP), such as Large Language Models (LLMs), can be employed to foster the usage and facilitate the documentation of Indigenous languages which are in danger of disappearing. %Although LLMs often require large amounts of data, w
%We show it is possible to fine-tune existing, pre-trained LLMs with tiny amounts of data to create language tools for such languages. 
We start by discussing the decreasing diversity of languages in the world and how working with Indigenous languages poses unique ethical challenges for AI and NLP. To address those challenges, we propose an alternative development AI cycle based on community engagement and usage. Then, we report encouraging results in the development of high-quality machine learning translators for Indigenous languages by fine-tuning state-of-the-art (SOTA) translators with tiny amounts of data and discuss how to avoid some common pitfalls in the process. We also present prototypes we have built in projects done in 2023 and 2024 with Indigenous communities in Brazil, aimed at facilitating writing, and discuss the development of Indigenous Language Models (ILMs) as a replicable and scalable way to create spell-checkers, next-word predictors, and similar tools. Finally, we discuss how we envision a future for language documentation where dying languages are  preserved as interactive language models. 
\end{abstract}

\section{Introduction}

Most of the recent extraordinary developments in Artificial Intelligence (AI) and Natural Language Processing (NLP), such as Large Language Models (LLMs), have predominantly used English language texts and data gathered in developed countries. These advancements have primarily targeted the needs and problems of those populations. Even within these countries, racial, ethnic, and linguistic minorities have been largely underrepresented in the construction of such models and technologies.

This paper describes research and work with Indigenous communities performed in the context of a joint project by \textit{IBM Research} and the \textit{University of São Paulo}, covering a period from early 2022 to mid-2024, under the auspices of the \textit{Center for Artificial Intelligence (C4AI)}\footnote{\url{https://c4ai.inova.usp.br/research_2/\#ProIndL_B_eng}.}.

The work described in this paper is premised on the need to increase the diversity of representation and knowledge in the technologies and language models being built. This involves encompassing a broad and diverse range of languages, peoples, places, and genders, as part of a social justice and decolonial agenda~\cite{buccella2023}.

In particular, we have been working for the past two~years to create AI technologies for Indigenous peoples in Brazil, targeting small communities in where Indigenous languages are still in use but under threat. About 200~languages are spoken currently in Brazil by between one to two million people\footnote{According to the 2010 and 2022 Brazilian Census, respectively.}, but the vast majority of these languages are in danger of disappear until the end of the century~\cite{moseley2010atlas}. Many of these languages are spoken by fewer than 100~people, often elderly, and are at immediate risk. Even the most spoken Indigenous language in Brazil, \textit{Tikuna}, probably has at most 50,000 speakers.

The projects and ideas described here explore the development of technologies to support Indigenous communities in documenting, preserving, and vitalizing their languages. Developing language technologies, both for speech and text, for these languages has been difficult in the past because of a lack of resources and linguistic knowledge and of appropriate computational technologies capable of working with small amounts of data. 

However, as discussed in a recent UNESCO publication~\cite{llanes2023digital}, \textit{``... artificial intelligence, natural language processing and automated speech recognition and voice processing could provide a crucial boost to language revitalization efforts, but these technologies need to be developed in accordance with the rights and provisions set by, among others, the Universal Declaration on the Rights of Indigenous Peoples and the principles of Indigenous Data Sovereignty [...]. Yet, the need to develop voice recognition, machine translation, speech processing, and text analysis technologies for Indigenous languages cannot be overstated.''}~\cite[p. 168]{llanes2023digital}.

The emergence of LLMs in recent years has positively changed the landscape of opportunities, in our view, for those efforts. Paradoxically, while LLMs need to be trained with gruesome amounts of data, at extremely high costs, they also have the ability to address new needs and requirements by methods such as \textit{prompt engineering}, \textit{Retrieval-Augmented Generation (RAG)}, and by \textit{fine-tuning} them with small amounts of data. Although the first two techniques do not help much in the context of endangered Indigenous languages, since it is unlikely that standard LLM models have ever seen samples of them, the fine-tuning approach has led to some cases of success (see a detailed review of related works in  section~\ref{sec:related_work}). 

\begin{figure*}[t!]
     \centering
    %  \includegraphics[width= 7.5 cm]{figures/america_languages_histogram.png} 
    %  \hspace{2mm}
    % \includegraphics[width= 7.5 cm]{figures/brazil_languages_histogram.png} 
    \includegraphics[width=7.8cm]{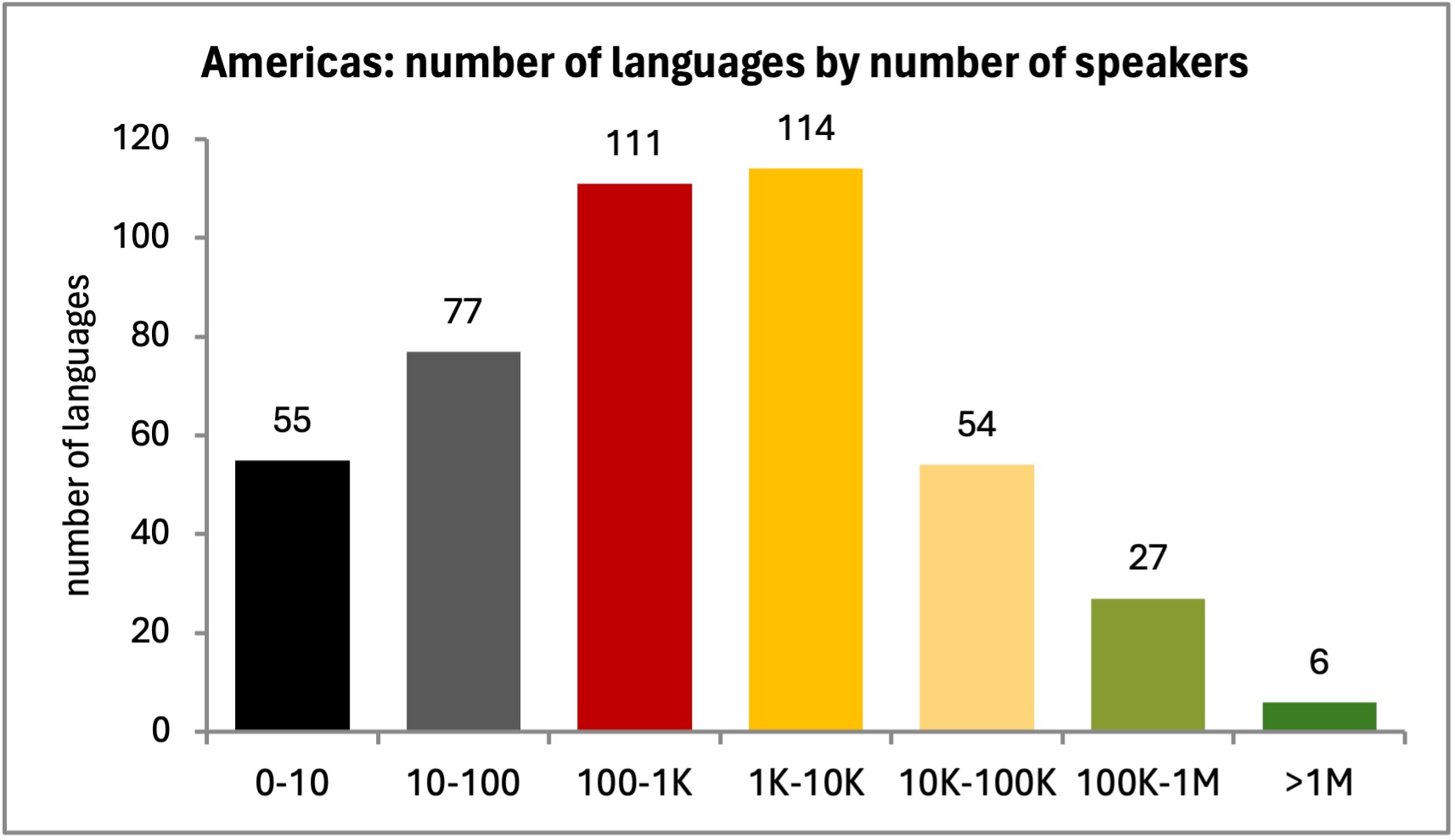} 
    %\\ \vspace{2mm}
    \hspace{2mm}
        \includegraphics[width=7.8cm]{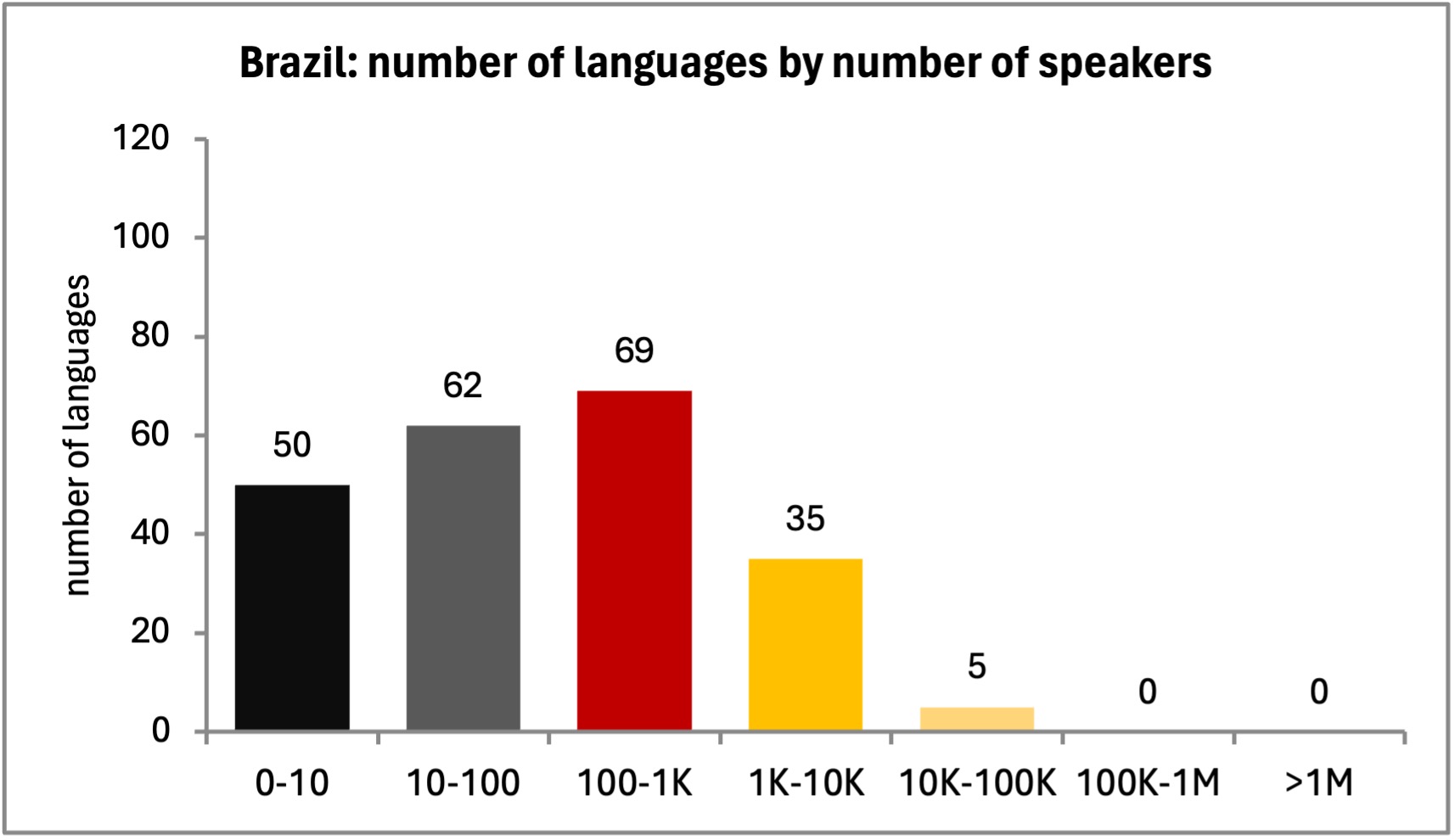} 
        
     \caption{Histograms of 444 Indigenous languages used in the Americas today and of 221 Indigenous languages in Brazil (based on 2010 numbers) with the number of languages for different logarithmic intervals of number of speakers.}
    \label{fig:america-languages-histogram}
\end{figure*}

Fine-tuning LLMs with tiny amounts of data is the main method explored in this work to create useful language tools such as automatic translators, spell-checkers, and editing tools. Some of the techniques described here may also applicable to domains outside Indigenous languages. Moreover, from a scientific standpoint, working with endangered Indigenous languages spoken by small groups of people has the additional benefit of guarantying that all fine-tuning effects come from this process and not from contamination errors, as discussed in~\cite{pinhanez2023balancing}.

The recent success of LLMs is, from the technical side, a strong incentive to work on text-based tools. However, orality is a major component of most Indigenous languages and, even more troublesome,  most of those languages use writing systems which were adapted from western languages~\cite{franchetto2020lingua}. In spite of that, creating tools to foster reading and writing is very important to strengthen endangered Indigenous languages. While young children are often the focus of vitalization efforts, it is common for Indigenous people to lose interest in their own language during the teenage to early adulthood period, by means of \textit{language attrition}~\cite{schmid2011contact}. 

The loss of interest in the native language is likely compounded by the interest in using and exploring the digital world at those ages. The challenge is furthered by the fact that it is generally easier to forget a language that one does not know how to read and write in~\cite{jayasuriya1992ethnicity,burn2014study}. We thus decided to focus our activities on creating writing tools and translators for endangered Indigenous languages, particularly for young people and written content creators of those languages. 

The works and ideas described in this paper are structured around three basic research themes. First, we have been exploring how to adapt current (and possibly create new) AI and NLP algorithms given the constraints imposed by ultra low-resource and endangered languages. Second, we have been investigating whether these AI- and NLP-based technologies can actually support Indigenous communities' language vitalization and documentation efforts, especially in the case of endangered languages. And third, we have been looking into how to promote the development and use of language technologies for Indigenous communities in a sustainable and ethical way.

\section{Language Diversity is Decreasing}\label{sec:decreasing_diversity}

Languages are the most comprehensive record of human linguistic and cognitive evolution~\cite{hale1992endangered} and documenting and analyzing them is as important as Archaeology and Anthropology for understanding humanity’s past. Moreover,
languages have distinct ways of organizing thinking and comprehending reality and society~\cite{harrison2008languages}. The disappearance of a language, when its last speaker dies, is equivalent to the destruction of an archaeological site or the extinction of a specie.

There are about 7,000 different languages spoken in the world today\footnote{\url{https://en.wal.unesco.org/world-atlas-languages}.}, with 4,000 of them spoken solely by approximately 370 million Indigenous people\footnote{\url{https://www.un.org/esa/socdev/unpfii/documents/5session_factsheet1.pdf}.}. Of these, 2,680 are likely to disappear by the end of the century~\cite{moseley2010atlas}. This threat to Indigenous languages has led the United Nations to establish 2022-2032 as the Decade of Indigenous Languages, in an effort led by UNESCO~\cite{unescodecade20}.
Language endangerment is a continuum, highly dependent on the number of speakers and, particularly, whether young children and teenagers speak the language.

There are multiple ways to prevent the disappearance of languages. These include ensuring they are taught and used in schools from  an early age, fostering their use by teenagers and young adults, promoting literacy for languages with a written form,  and making them official languages in regions with significant numbers of speakers. Additionally, it is important to provide social and government services  in those languages and to encourage  their use in digital contexts and social media. However, the appropriateness of different policies and tools  depends on many factors including  the number and age profile of the language's speakers.

As an example of the diversity of endangerment conditions among Indigenous language, consider that there are about 1,000 languages used today in the Americas\footnote{\url{https://en.wikipedia.org/wiki/Indigenous_languages_of_the_Americas}.}. We could found data on the number of speakers for 444 of these languages\footnote{Combining data from \url{https://en.wikipedia.org/wiki/Indigenous_languages_of_the_Americas}; \url{https://www150.statcan.gc.ca/t1/tbl1/en/tv.action?pid=9810027101}; \url{https://www.ibge.gov.br/en/statistics/social/population/18391-2010-population-census.html?edicao=19316&t=resultados}, table 1.13; \url{https://www2.census.gov/library/publications/2011/acs/acsbr10-10.pdf}.}.
Figure~\ref{fig:america-languages-histogram} shows the histograms of the number of languages considering different logarithmic intervals in the number of speakers for those 444 American languages and for 221~languages spoken in Brazil. 

In the Americas, there are 6~languages spoken by more than 1 million people, notably \textit{Guarani}, with 6.5 million speakers, and \textit{Southern Quechua}, with 5 million speakers (as a reference, a number comparable to the speakers of \textit{Bulgarian} in the world). This group, together with the 27~languages with more than 100K speakers, such as \textit{Yucatec Mayan}~(890K), \textit{Mapuche}~(260K), and \textit{Navajo}~(170K), are thriving languages, with reasonable support from digital tools such as automatic translators and word editors. There is no Indigenous language in Brazil with a similar number of speakers.

The next interval consists of 54~languages with between 10K and 100K speakers, including languages such as \textit{Cree}~(96K), \textit{Tikuna}~(47K), and \textit{Nheengatu}~(20K), which already exhibit signs of endangerment, such as a limited number of young speakers, and have almost no support from digital tools. This group includes 5~languages spoken in Brazil. The subsequent interval comprises 114~languages with between 1K and 10K speakers, such as \textit{Choctaw}~(9.6K), \textit{Guarani Mbya}~(6K), and \textit{Mohawk}~(3.9K), of which about 35~are Brazilian languages.

\begin{figure*}[t!]
     \centering
     \includegraphics[width=7.5cm]{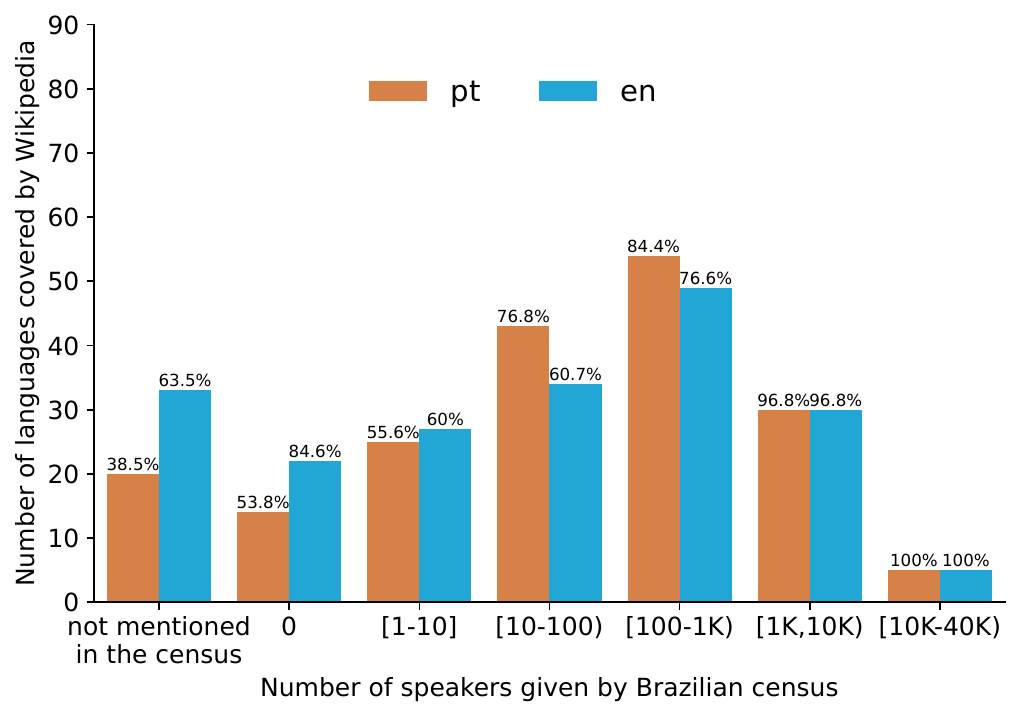} 
     %\\ \vspace{2mm}
     \hspace{2mm}
     \includegraphics[width=7.5cm]{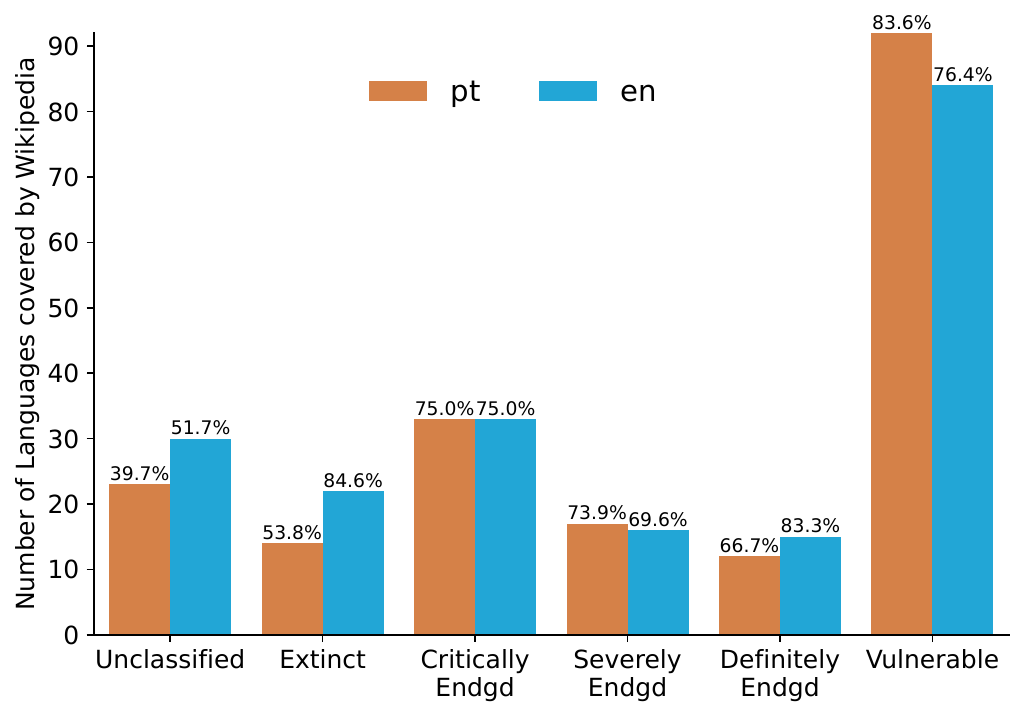} 
     \caption{Histograms of the number of Brazilian Indigenous languages with a descriptive page in Wikipedia per number of speakers (left) and according to different levels of endangerment (right).}
    \label{fig:descriptiveWikipedia}
\end{figure*}

We consider these two groups of 169~languages (marked in orange in figure~\ref{fig:america-languages-histogram}) as the major target for the research described here. For these languages, there is enough written data, as shown later, to allow the development of digital text tools such as translators, and enough digital and social media literacy to enable their vitalization through digital means. Notable examples are the two~languages we have been working with, \textit{Nheengatu} (20K) and \textit{Guarani Mbya} (6K). 

The interval between 100 and 1,000 speakers, with 111~languages in the Americas, marked in red in figure~\ref{fig:america-languages-histogram}, includes languages such as \textit{Chocho}~(810), \textit{Nambikwara}~(573), and \textit{Koyukon}~(300). These languages are in severe danger of disappearing and are more likely to benefit from the use of digital technology as a support for documentation and analysis. They are also good candidates for the creation of interactive language preservation models described at the end of this paper. Of these languages, 69~are spoken in Brazil.

Finally, languages with up to 100~speakers, marked in grey and black in figure~\ref{fig:america-languages-histogram}, are in severe danger of disappearing within 20~years or less, as most of their speakers are elderly. Traditional efforts of documentation and preservation, possibly aided by digital means, seem to be the most applicable methodology here. There are about 132~languages in these conditions in the Americas, with 112~being languages spoken in Brazil. However, the numbers of both languages and speakers for this range are notoriously imprecise.

The digital situation of languages with fewer than 1,000 speakers is often bleaker. In a study performed in the context of our project~\cite{vasconcelos2024}, we found that many Indigenous languages spoken in Brazil are likely to disappear without even leaving essential digital traces. Figure~\ref{fig:descriptiveWikipedia} ~(left) shows that about half of the languages with 10~speakers or fewer have no entries in Wikipedia about their existence, either in English or in Portuguese. Similar observations were made for about 20\% of the languages in the 10 to 1K range. Beyond 1,000~speakers, virtually all languages have a descriptive entry. However, as discussed by~\citet{vasconcelos2024}, many of these pages have limited quality and content, or have a reduced number of active editors, so information is likely to be limited and outdated.  Moreover, the lack of entries correlates with the level of endangerment of the languages, as shown in figure~\ref{fig:descriptiveWikipedia}~(right).

\section{Working with Indigenous Communities}

Our work has been guided by the principle that technologies and solutions for Indigenous peoples must be developed with them, as stated in the \textit{Declaration of Los Pinos}: \textit{“Nothing for us without us.”}~\cite{lospinos2020}. 

Since the start of this project in 2022, we have been engaging with many communities in Brazil, including the \textit{Guarani Mbya}, \textit{Guarani Kaiowá}, \textit{Guarani Nhandewa}, \textit{Tupi}, \textit{Terena}, \textit{Baré}, \textit{Wassu}, \textit{Tukano}, \textit{Pankararu}, \textit{Zoé}, and \textit{Mehinako} peoples. We have also engaged organizations which work with Indigenous communities in Brazil, such as the Ministry of Indigenous Peoples, the Interamerican Development Bank (BID), the Federation of the Indigenous Organizations of the Rio Negro (FOIRN), the National Foundation of Indigenous Peoples (FUNAI), the Plurinational Union of Indigenous Students (UPEI), the Socio-Environmental Institute (ISA), and many other NGOs. 

These engagements have resulted in two projects with communities of speakers of two languages, \textit{Guarani Mbya} and \textit{Nheengatu}, which we describe in this section. However, before discussing the projects, it is important to discuss the ethical principles and guidelines we have used to frame our engagement with those communities.

\subsection{AI Research with Indigenous Peoples}\label{subsec:ai_research_indigenous}

Doing research with Indigenous peoples is subjected to specific guidelines and legal constraints. \citet{mihesuah1993suggested} is a good example of a comprehensive set of guidelines for research with US American Indigenous communities. 
\citet{straits2012guiding} also proposed a set of guidelines on how to engage in research with Native US American communities based on 11~principles, including native-centrism, co-learning and ownership, continual dialogue, transparency and accountability, integrity, and community relevance. 

Most importantly, there is a lot of distrust from many Indigenous communities towards researchers and academic work, resulting from a history of exploitation, disregard, and knowledge extractivism: \textit{``... [the word research] is probably one of the dirtiest words in the Indigenous world’s vocabulary.''}~\cite[p.~1]{smithDecolonizingMethodologiesResearch1999}.
To address these issues, \citet{smithDecolonizingMethodologiesResearch1999} proposes, as part of a decolonization perspective, that \textit{relational accountability}~\cite{wilsonResearchCeremonyIndigenous2008}  should guide these engagements since it is inherent to Indigenous ways of doing. The key ideas behind relational accountability is that relationships with the communities are important in research and that all the parties are responsible for maintaining them.

Besides  ethical considerations, there are  legal and regulatory procedures that must be followed in different countries when working with specific Indigenous communities. Particularly relevant to our cases, of developing AI-based tools, are issues related to data sovereignty, consent, and intellectual property (IP) rights~\cite{harding2012conducting} which should be considered in tribal research in the USA. Research, in such cases, should include  special procedures for informed consent processes and the involvement of community members in defining exposure and risk to the community. Similarly, \citet{sahota2007research} discusses the need for research regulation in American Indian and Alaska Native communities and the challenges to establish this regulation. 

%In the case of research specifically related to the digital realm, 
\citet{llanes2023digital} has proposed a research framework based on 7~key approaches for digital initiatives with Indigenous languages, based on: facilitating digital communication in Indigenous languages; multiplying Indigenous language content online; normalizing the use of Indigenous languages online; educating in and teaching Indigenous languages online; reclaiming and revitalizing Indigenous languages and knowledge digitally; imagining and creating new digital media in Indigenous languages; defending spaces for Indigenous languages and linguistic rights; and protecting Indigenous linguistic heritage and communities. 

The research described in this paper aims to explore and develop technologies which are related to many of those issues and, in particular, to the need to normalize the usage of Indigenous languages in the online space. Similar issues are also discussed by~\citet{oliveiraetal_helloindigenous2024}, with a focus on practical aspects such as fonts, keyboards, and Unicode representations.

When it comes to AI- and NLP-related research, gathering, controlling, and using linguistic data become essential parts of the research process and, therefore, of ethical concerns. The sovereignty of Indigenous data is an area that has attracted considerable discussion in recent years, particularly in the case of population and genetic data~\cite{kukutai2016indigenous,walter2021indigenous,kukutai2023indigenous}. Its importance for digital and AI research has attracted some recent attention~\cite{llanes2023digital}. There are also some efforts to create new agreements and practices, such as the \textit{Kaitiakitanga} data license\footnote{\url{https://github.com/TeHikuMedia/Kaitiakitanga-License}} proposed by members of the \textit{Maori} language community (see also the various links provided by \textit{TeHiku}\footnote{\url{https://tehiku.nz/te-hiku-tech/te-hiku-dev-korero/25141/data-sovereignty-and-the-kaitiakitanga-license}.}).

A more specific guide for AI-related work, including tool and technology design methodologies, was proposed by the \textit{The Indigenous Protocol and Artificial Intelligence (A.I.) Working Group}~\cite{lewis2020indigenous} as a result of two workshops with Indigenous leaderships, linguistic professionals, and computer researchers. Nevertheless, although large AI conferences have hosted workshops dedicated to Indigenous contexts, the discussion of ethical guidelines when working with Indigenous peoples is still limited within the AI community.

The engagement principles and ethical guidelines used in our research have been discussed in more detail in ~\cite{pinhanez2023balancing}. In particular, the paper describes a practice, which we have adopted in our work as described in the next sections, of putting in place \textit{damage containment procedures} when dealing with Indigenous data.

Such damage containment procedures are based in four different damage mitigation actions~\cite{pinhanez2023balancing}. First, we ensure that everyone involved in handling language data is fully aware of the ethical issues and the dangers of releasing data without the proper authorizations. 
Second, as part of the containment process, we do not release the data or the created models publicly, not even among other AI researchers, which is a common practice in the field.
Third, we make sure that prototypes of tools, especially in actual systems, are only deployed or tested with express authorization from the community, as they may generate inappropriate or offensive language. Fourth, the existence of the data and its associated models, including their shortcomings, is transparently disclosed to the Indigenous communities and other stakeholders, and protocols to control and use the data are established.

\subsection{Working in a Guarani Mbya School}\label{subsec:working_mbya}

In the case of the works described in this paper, we started in 2022 to contact various Indigenous groups and organizations in Brazil. Following a series of meetings with the \textit{Tenondé Porã} community on the outskirts of São Paulo city, in 2023 we were invited by the community to explore the use of writing assistants by Indigenous high school students and to conduct activities fostering community-lead linguistic documentation and analysis. This school is part of an Indigenous land, home to approximately 1,500 people,  where the main language is \textit{Guarani Mbya}. %At that time, this was one of the few Indigenous areas in Brazil open to this kind of work due to the restrictions imposed by the COVID-19 pandemic.

The \emph{Guarani Mbya} language is spoken by approximately 6,000 people in Brazil~\cite{morelloseiffertinventariombya11}, mostly in the South-Southeast region. Although it is   still actively spoken and well-studied, there are few sources of translated texts and digitized data. Guarani Mbya belongs to the \textit{Tupi} linguistic family and it is related to the \textit{Guarani} language spoken by millions of people in Paraguay and Bolivia. However, it is as different from Guarani as Portuguese is from Spanish.

\begin{figure*}[t!]
     \centering
     \includegraphics[width=7.8cm]{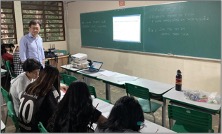}
     \hspace{2mm}
      \includegraphics[width=7.8cm ]{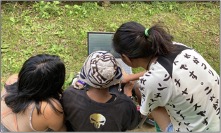} 
     \caption{Photographs taken during the workshops at the Gwyra Pepo Indigenous High School in 2023.}
    \label{fig:tenondephotos}
\end{figure*}

The invitation from the Tenondé-Porã community led to weekly 2-hour workshops where various technologies and prototypes were introduced to and used by the high school students of the \textit{Gwyra Pepo Indigenous Guarani School}. The students explored these technologies and discussed how to modify and improve them. As a result of this work, we developed an initial prototype of a writing assistant for the Guarani Mbya language, which included rudimentary electronic dictionaries, and word prediction, and basic translation. 

The technical aspects of those tools are discussed in sections~\ref{sec:translators} and~\ref{sec:writing-assistant} but some of them could only be built because we applied state-of-the-art AI technology. For instance, the translator was developed by fine-tuning a German-English high-quality translator using data obtained from traditional dictionaries, schoolbooks, and folk tales, mostly from the collection of Prof. Robert Dooley~\cite{dooley1985,dooley1988a,dooley1988b,dooley2016}, an expert in Guarani Mbya.
%\marisa{Marisa: Footnote perdido?}

We faced a major challenge when we began designing the workshops with the high school students. As discussed in more detail in~\cite{cuichi2023paper}, the last 15~years have produced many scholarly works with cases of engagement with Indigenous communities for the design of digital artifacts~\cite{kagonyaetal_techinsightskenyandiaspora_chi15,munteanetal_designingculturalvalues_chi17,tayloretal_situationalwhen_chi17,muashekeleetal_appropriationtechindigenous_ct19,reitsmaetal_respectfuldesignindigenous_2019,tzouetal_storyworkinSTEM-Art_2019,moradietal_italarchive_nordhi2020,shedlockhudson_maoriItartefacts_2022}. Most of these works focus  describing key principles that designers should follow when engaging with Indigenous communities. %in ``appropriate ways''. 
%For instance, \cite{reitsmaetal_respectfuldesignindigenous_2019} describes the process of co-designing interactive artworks with Sarawak communities in Malaysia based on the \emph{respectful design} concepts of~\cite{sheehan_respectfulDesign_2011}.

However, while this literature offer extensive descriptions of \textbf{what} should be done in the design process, it often provides very limited on \textbf{how} the actual design process should happen. For instance,  when considering the use of \emph{Respectful Design}, proposed by~\citet{sheehan_respectfulDesign_2011}, we could not find any literature on how it should be implemented in actual design processes. %Showing respect can be done in many different ways, and there were almost no examples of how that can be done in complex multi-cultural settings like a design workshop.
%What we found was a possible gap in design knowledge about specific methods to engage meaningfully, respectfully, and appropriately with an Indigenous community. 
There are a few exceptions were details about how design workshops were  conducted were actually provided~\cite{tayloretal_situationalwhen_chi17,muashekeleetal_appropriationtechindigenous_ct19,tzouetal_storyworkinSTEM-Art_2019}. However, even in those cases a systematic description of the  methods and means was never provided nor 
good ideas and practices on how to actually conduct the design workshops. %\marisa{Marisa: tem um ponto final entre provided e good ideas.}

Interestingly, some works discussing community research related to Indigenous education, schools, and connectivity~\cite{franchetto_guerraanalfabetos_2008,hermesetal_designing-indigenous-language-revitalization_2012, arola_indigenousinterfaces_2017,leal_connectivity_2022}
and Indigenous writing~\cite{johnson_writinglessonrevisited_1997,franchetto_guerraanalfabetos_2008} 
were more helpful. %the ones providing some inspirations and ideas for our initial structure of the design workshops. 
In particular, the discussion in~\cite{arola_indigenousinterfaces_2017} about \textbf{being} an Indigenous person vs. \textbf{doing} the Indigenous way was enlightening and helped us to outline some initial ideas for the workshop.

Based on this concept, we structured our workshops around three principles. First, we adopted the conversational structure observed in the three meetings we had with the leadership of the Tenondé-Porã community. In those meetings, everyone was welcomed to voice their opinions, whenever they felt comfortable, and the discussion progressed at its own pace toward a consensus. %\marisa{ponto final entre consensus e humor}
In our work at the high school, students could enter and leave the classroom, stop to smoke their pipes, 
small children were welcomed to observe, 
and humor and laughing were valued and welcomed, similar to what we had observed in the meetings with the community. 

Second, we began the process by exploring the students' current activities on the Internet, particularly on \emph{WhatsApp} and in the game \emph{Free Fire}. This served as a starting point to examined their use of written language and how a language aid could play a role in their current usage. For instance, students reported to us that sometimes they used their native language as code to talk among themselves with privacy in public forums of Free Fire.

Third, we tried to downplay our own eminence as scholars, linguists, designers, and computer wizards. On the second workshop, we presented a version of a Guarani Mbya to Portuguese translator which we knew had a poor performance. The goal was to showcase our incompetence in their language and also to present ourselves as people who can fail, sometimes miserably. 

In total, 14~workshops were conducted over three~months, during  which different versions of the writing assistant prototype and its components were used and discussed within the context of various writing activities. One of the best workshops happened when we used, as a writing theme,  images from a recent protest by the Guarani community in São Paulo for land rights and the subsequent confrontation with the local riot police. Possibly, since many of them knew people involved in the conflict, they felt more interested and engaged to write to document their struggles.
%\marisa{Marisa:mais detalhes do porquê foi interessante para eles.  }

This first engagement had limited outcomes in terms of creating high-quality technology or deploying actual writing tools. Instead, its main result was to provide a context for us, for the students, and for the community to understand the opportunities and challenges involved in bringing AI to the realm of Indigenous languages. 

For the researchers involved in the workshops, this engagement demonstrated the need for good writing tools and methods to support a generation of students who, despite being fluent in their native language, were still learning how to write in it. While these youngsters were actively involved in writing messages among themselves, reading social media, and sharing content, the presence of Guarani Mbya text in their virtual lives seemed to be almost nonexistent.

We concluded that there is a pressing need to develop tools to support writing among youngsters, who encounter difficulties translating concepts and ideas they can easily express verbally into text. Also, it became evident how diverse their writing abilities were and the challenge of creating supporting tools that could be used by students of different levels of literacy.

Conversations with the students during and after the workshops led us to believe that the workshops had some positive impacts on the participating students. The presence of a team of individuals, clearly unable to understand their language but interested in developing tools to enable them to write in it, was recognized by some as an indication of the value placed on their culture. We also observed some students expressing interesting on learning how to use computers to improve their writing skills, on how to employ them to gain a better understanding of their own language, and even on how to program them. 

Finally, for the community where the school is located, our project seemed to have contributed to a larger discussion about how the community wants to interact with the digital world and the Internet. This is a community where high-speed Internet had arrived just one year before our engagement, through an optical cable provider, since before there was very poor cell phone coverage in the area. When we engaged with them, there were ongoing debates about which websites and apps the community would allow access to, and when. 

\begin{figure*}[t!]
    \begin{subfigure}{0.38\textwidth}
        \includegraphics[width=\linewidth]{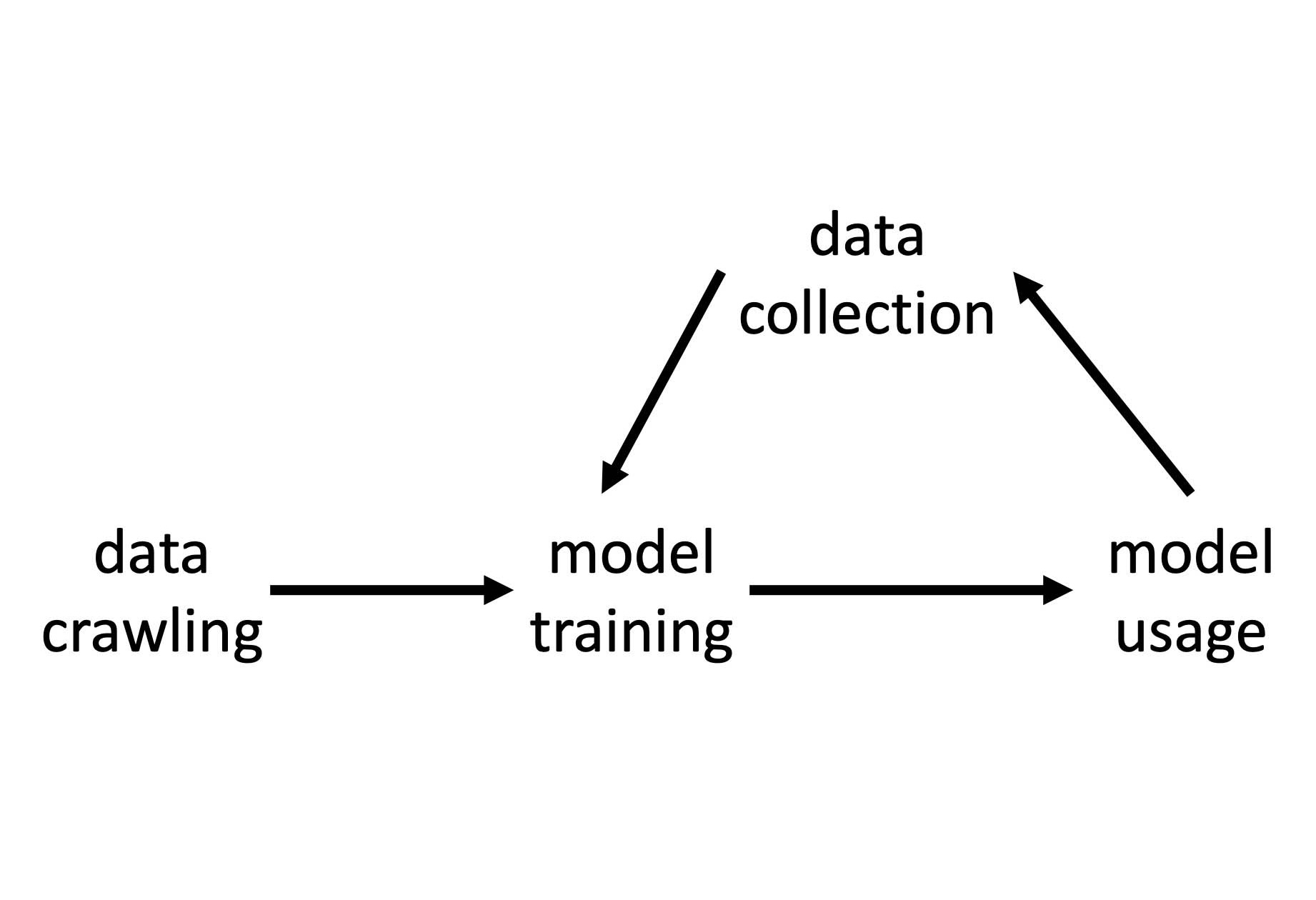}
        \caption{Traditional AI development cycle.}
    \end{subfigure} 
   \hfill
    \begin{subfigure}{0.58\textwidth}
        \includegraphics[width=\linewidth]{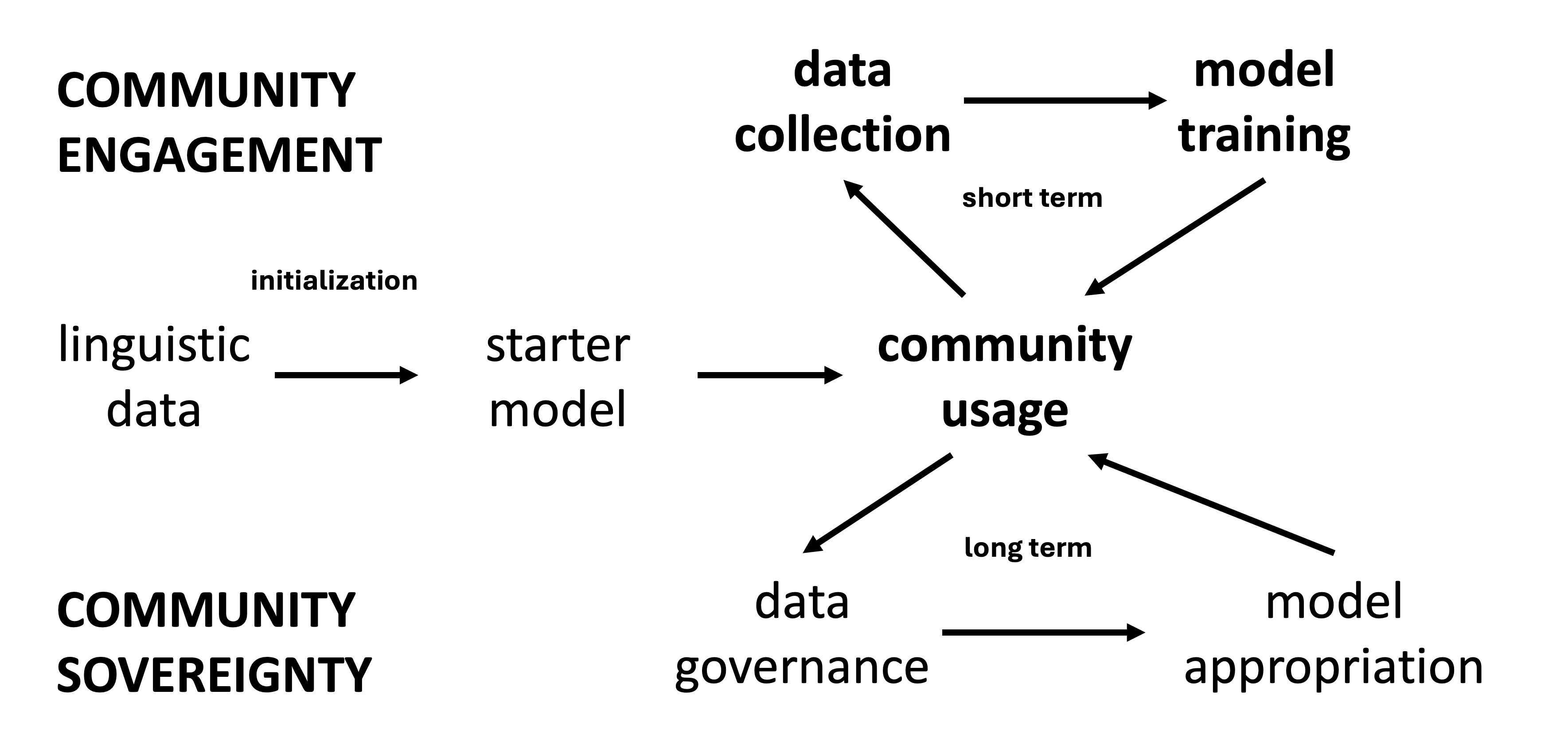}       
        \caption{AI development cycle for Indigenous communities.}
    \end{subfigure}
    % \hfill
    % \begin{subfigure}{0.49\textwidth}
    %     \includegraphics[width=\linewidth]{figures/writing_assistant_translation_border.png}
    %     \caption{translation}
    % \end{subfigure}

    \caption{Traditional AI development cycle (a) and the proposed AI development cycle for Indigenous communities with emphasis on community usage, engagement, and sovereignty (b).}
    \label{fig:ai_cycles}
\end{figure*}

We believe our project served as an opportunity for the community to explore potentially positive uses of these resources, in a controlled experiment. In fact, after the four~months of the project, the Tenondé-Porã expressed concerns about exposing their language and culture to the Internet. They requested that our activities be suspended while the community deliberated on the proper continuation of the project. We assured them that no language data had been collected during the process, thanked them for the opportunity to collaborate with their students, and suspended further work on the Guarani Mbya writing assistant, accordingly.

\subsection{An AI Development Cycle Suitable for Indigenous Communities}\label{subsec:ai_cycles}

Consider all the different ethical issues discussed in sections~\ref{subsec:ai_research_indigenous} and the lessons learned working with the Guarani Mbya students explored in~\ref{subsec:working_mbya}, we started to question whether the traditional ways AI has been often developed would work for Indigenous communities. 

Figure~\ref{fig:ai_cycles}.a shows the traditional AI development cycle to text-based systems which starts with an initial cycle of data gathering, often from the Internet, identified as \textit{data crawling}, followed by \textit{model training}, and a repeating cycle of \textit{model usage},  \textit{data collection} from the users' interaction with the model, and retraining of the model.

Is Internet crawling a feasible strategy for endangered Indigenous languages? Putting aside for a moment the issue of how to get approval and permission from Indigenous peoples to use data obtained from the Internet, an important technical challenge is to find public websites containing text in endangered Indigenous languages. As discussed before, data from disappearing languages is scarce~\cite{vasconcelos2024}, even for the languages  considered here, with the number of speakers ranging from  1,000 to 100,000. 

To find this data, which is dispersed among vast amounts of content in more used languages, it is fundamental to employ automatic mechanisms for \textit{language identification}. In~\cite{cavalin2023understanding} we explored the development of language identifiers for 7~Brazilian Indigenous languages, using an \textit{SVM classifier} with \textit{bag-of-words} features trained on data from the Bibles dataset described in section~\ref{sec:translators}. When tested with non-Bible data, recognition accuracy varied from $37.3\%$ to $91.7\%$ in test datasets with only Indigenous languages samples. Accuracy was higher when similar languages from the same linguistic family were considered. However, in a realistic evaluation where the data from more commonly used non-Indigenous languages were also present in high proportions, the performance was much worse~\cite{cavalin2023understanding}.

The first impulse of AI developers and scientists in such conditions would be to engage the community to create a data gathering mechanism based on the promise of future use of the system. In our experience, this is very difficult because of both the difficulties in communicating how AI is built and the common distrust of those communities towards research projects as discussed before. Instead, our experience with the Guarani Mbya people suggested that a different framework for AI development is needed, centralized on the use of the tools by the community.

Figure~\ref{fig:ai_cycles}.b shows a diagram of the proposed AI development cycle. First, it is anchored on \textit{community engagement} and~\textit{community sovereignity}, as central pillars to ethically and sustainably realize the focus of the development which is \textit{community usage}. To enable the initial use by the community, there is an \textit{initialization} step, with approval of the community, where a \textit{starter model} is built based on \textit{linguistic data} such as publicly available dictionaries, theses, grammars, books, and similar materials. The starter model is the seed for community usage, even if it has shortcomings such as the translator we used with the Tenondé Porã students.

In the AI development cycle we are proposing, there are two sub-cycles associated the community usage, \textit{short-} and \textit{long-term}. The short-term sub-cycle follows the path of traditional AI development with data collection and model training activities, but in this case with direct participation of the community in key decisions and under its strict supervision. Also, all data collected is considered private, belongs to the community, and is used only when authorized. The long-term sub-cycle refers to the establishment of structures and processes of \textit{data governance} and to the training of people and the creation of mechanisms to allow \textit{model appropriation}.

The centrality of \textit{community usage} is key because it guarantees both that the tools being developed are useful and used and that the research and development efforts provide a clear and immediate return to the community. However, we still have to demonstrate that development of AI tools and systems based on this framework is feasible, useful, and efficient. Some initial works exploring the framework are presented in the next sections but first we describe our engagement efforts to find a community to explore those ideas with.

\subsection{Working with Nheengatu Speakers}

%The work with the Tenondé-Porã students made clear for us the importance of writing tools as a way to engage Indigenous teenagers and young adults in using their language. 

Following our engagement with the Guaranis, we decided to focus on another Indigenous language, \textit{Nheengatu}, which is spoken by approximately 20,000 people across three different areas of the Amazon area and in the Northeast of Brazil. %\textbf{TO COMPLETE: more details to be added}. 
This language is used by various peoples and ethnicities, including cases where the language was adopted by groups after the loss of their original language, such as the \textit{Baré} people~\cite{epps2013upper}. 

The choice of Nheengatu was motivated the availability of public data and knowledge, its  multi-ethnic characteristic, the participation of linguistic experts at the University of São Paulo, and ongoing Indigenous-led initiatives focused on translation in Nheengatu. Also, there have been some recent efforts to bring Nheengatu to the digital world, such as the self-learning tool \textit{Nheengatu app}\footnote{Developed by Suellen Tobler, \url{https://heengatu-app.web.app}.}. 
%\textbf{TO COMPLETE: Nheengatu app, computational linguistics, translations, etc.}.

In our on-going engagement with the community, we have been working on two basic workstreams. %\marisa{tiraria essas palavras de guerra, talvez facets.}. Firstly, similar to our work with the Guarani Mbya community, we are focused on the \textbf{development of writing assistants} to foster language use in textual form.
First, we have recently started a partnership with a group of Indigenous students, mostly from the Baré ethnicity, from the State University of Campinas. %They are looking for easier methods to write in their own language on computers and mobile phones. 
This collaboration will involve a structured writing workshop using a new and improved version of the Nheengatu writing assistant, ultimately leading to the development of educational and cultural materials. %One of the Indigenous collaborators is also the leader of an educational project in Manaus, the largest city of the Amazon area, and we are exploring the idea of using the writing assistant to support the alphabetization process of children.

%\marisa{replace front. Estou trocando algumas palavras problematicas, tipo explore tambem. }
The second workstream aligns with the need for more comprehensive translation services and materials. The availability of translation services, especially from dominant to Indigenous languages, is important to increase the accessibility to legal, health, schooling, and other government services.  At the same time, it is important that Indigenous individuals are able to communicate in their own language to denounce crimes, request services, and participate in the political and social debates, making the translation from their languages to a dominant language also essential to citizenship. 

In both scenarios, the goal is not to replace the Indigenous translators but rather to make their work more productive and reliable, allowing an often small number of translators to support a larger volume of demands and tasks. We are currently establishing a collaboration with a team of professional translators and writers in the Nheengatu language, who recently undertook the translation of the Brazilian Constitution into Nheengatu\footnote{\url{https://www.cnj.jus.br/wp-content/uploads/2023/07/constituicao-nheengatu-web.pdf}.}. The goal is to collaborate with them in developing a version of the writing assistant tailored for translation tasks and, possibly, focusing on legal and healthcare texts, incorporating as much as possible high-quality automatic translators as a way to enhance both productivity and quality.

Our plan is to work with those communities to build AI-based tools for Nheengatu as a proof point that it is possible to develop such tools for  endangered Indigenous languages, in their typical contexts of ultra-low data resources. In the next two sections, we present the methods and technologies we have been exploring to create prototype tools together with Indigenous communities, following the %ethical guidelines we have just discussed and the 
proposed community-based AI development cycle depicted in  figure~\ref{fig:ai_cycles}.b.

%%%%%%%%%%%%%%%%%%%%%%%%%%%%%%%

\section{Building Translators with Ultra-Low Amounts of Linguistic Data}\label{sec:translators}

%As discussed before, our work has been focusing on developing technology along two workstreams: digital writing assistants for young people and translation services. 

In this section we look into the technical challenges and the solutions we have found to create bilingual \textit{machine translators (MTs)} for Indigenous languages based solely on publicly available linguistic data, following the \textit{initialization} process of the development cycle of figure~\ref{fig:ai_cycles}.b. Given the extremely limited amounts of data available for most Indigenous languages, especially for endangered ones, the development of translators for such languages is only feasible today due to recent developments in AI technology, such as the use of \textit{Transformer} technologies~\cite{vaswani2017attention} and the availability of open pre-trained \textit{Large Language Models (LLMs)}~\cite{joo2023survey}. 

%Our team has focused not only on overcoming the associated technical challenges but also on making sure that the technology we create is useful and supports the needs of Indigenous peoples, what we are accomplishing by trying to work as much as possible directly with them.

The most common way to create MTs for low-resource languages such as endangered Indigenous Languages involves taking generic LLMs, pre-trained on large corpus using self-supervised techniques with high levels of data, and \textit{fine-tune} them with a much smaller parallel downstream corpus in the target language~\cite{lee-etal-2022-pre,mager-etal-2023-neural}. This usually results in better translation accuracy than training  from scratch with limited data~\cite{adelani-etal-2022-thousand}. Additionally,  some results suggest that translation quality can be improved by using data from multiple languages or multilingual models~\cite{saleh-etal-2021-multilingual-neural}.

However, our experience in building translators for Indigenous languages has led us into a different direction. First, %by analyzing the results of experiments creating bilingual and multilingual translators for 39~BILs, using Bible data, 
we saw that multilingual translators often achieve falsely improved accuracy results by adopting a ``cheating'' strategy of memorization~\cite{cavalinnaacl24}.
Furthermore, we investigated the impact of adding more fine-tuning data, particularly in the situation where the additional data raised  ethical concerns~\cite{dominguessigul24}. %This was done in the context of a Guarani Mbya to English translator, initially created using dictionary data, and later improved by fine-tuning with Bible data. As discussed before, the Bible is a controversial text for many Indigenous peoples, and, 
Unfortunately, we observed traces of contamination in the outputs, though in limited numbers and scope.
We also found that the most significant improvement in accuracy was obtained by manually enhancing the quality of the fine-tuning training data. % used to create bilingual translators. %These results were further validated  through a manual evaluation of the effectiveness of a translator for the Nheengatu language~\cite{dominguessigul24}, where we observed that over half of the outputs had a good quality, despite being trained with only about 6,200 pairs of sentences, extracted mostly from dictionaries, lexicons, and bilingual story- and textbooks. %\marisa{Nao sei se deveria explicar o que vcs querem dizer com quase perfeito.}

These results and associated challenges, discussed in technical detail next, suggest that it is feasible to fine-tune LLMs into valuable bilingual translators using data commonly available for Indigenous languages that have been reasonably documented and studied by linguists, if some key methodological conditions, such as data cleanliness, are met. We start by describing the data sources we used in those works and  ethical issues related to them.

\subsection{The Data Used in our Research}

To develop both writing assistants and translators, we have been exploring machine learning methods based on small amounts of data. %Nevertheless, there is still need of some data to train, fine-tune, or test the models we have been working on.
The data used in our work can be divided into two basic types: data extracted from \textbf{linguistic sources}, such as dictionaries, lexicons, theses, and publicly available books; and data extracted from \textbf{multiple versions of the Bible}  available on the Internet. 

We have been working exclusively with \textit{Brazilian Indigenous languages (BILs)}. Brazil was home to about 270 Indigenous languages according to the 2010 Census~\cite{ibgelinguasindigenas2010}, although some linguistic experts believe the actual number is more likely to be close to 200~\cite{franchetto2020lingua,storto2019}. Those languages were spoken by approximately 800,000 people~\cite{ibgelinguasindigenas2010}, with half living in Indigenous lands, although the 2022 Census revised this number to about 1.7~million\footnote{\url{https://biblioteca.ibge.gov.br/index.php/biblioteca-catalogo?view=detalhes&id=2102018}.}. \citet{storto2019} provides a good overview of the history, structure, and characteristics of a few BILs. Almost all of these languages are considered endangered~\cite{moseley2010atlas}, remnants of the 1,000 languages estimated to be in use in Brazil before the arrival of Westerners 500 years ago~\cite{Rodrigues_2019}.

%\marisa{Marisa: Aqui fiquei na duvida, se nao tem permissao, vcs coletam mesmo assim?}
The first type of data comprises data extracted %, with permission when possible, f
from publicly available dictionaries, lexicons, theses, and books. For the languages in the range of 1,000 to 100,000 speakers we are targeting, such sources are common, mostly created through linguistic research and educational efforts. We have mostly collected resources for two languages we have been working with, Guarani Mbya and Nheengatu. The data, often in the form of \textit{PDF} files, are processed by scripts and manually, resulting in datasets with pairs of sentences, for the training of translators, or well-structured sentences, for the training of encoders of Indigenous languages models and to be used in development of the writing-support tools. Following the previous discussion, we do not release publicly or share this data without the permission of the associated Indigenous communities.

The second type of data was collected in the early stages of the project by researchers from IBM Research and comprises 39~Indigenous languages spoken in Brazil, for which we found translations of the \emph{New Testament} of the \textit{Bible}, a book with about 7,000 verses in its English versions. The Bible is often available in many of these languages  due to translations by Christian churches~\cite{franchetto2008guerra}. Table~\ref{tab:39bils} lists these languages, including 36~spoken primarily in Brazil and 3~Guarani-related languages used mostly in Paraguay and Bolivia, but also spoken in some areas in Brazil. We adopted the Indigenous language classification, nomenclature, and data from the 2010 Brazilian Census by IBGE~\citep{ibgelinguasindigenas2010} and language acronyms according to ISO~639-3.

%% Edit table
\begin{table}[t!]
    \centering

    \includegraphics[width=7.6cm]{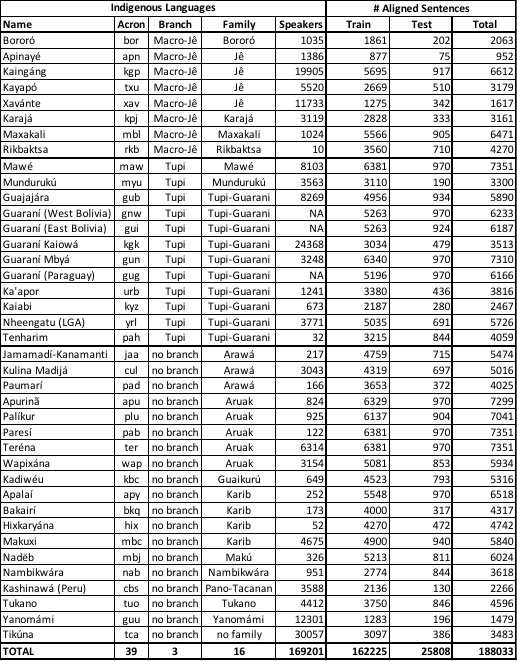}
    \caption{Indigenous languages and corresponding dataset sizes used in the study. 
    %Besides Brazilian languages, some languages from Bolivia, Paraguay, and Peru are also considered.
    Language name, branch, family, and number of speakers (considering only those who speak the language at home in an Indigenous land in Brazil) according to the table 1.13 of the Indigenous data of the Brazilian census of 2010~\cite{ibgelinguasindigenas2010}. %Number of aligned sentences considers the New Testament of The Bible data which is available both in Portuguese and English language.
    }
    \label{tab:39bils}
\end{table}

This dataset, referred as \emph{Bibles}, was mainly obtained from the \emph{ebible} website\footnote{\url{https://ebible.org/download.php}.}. A few other languages were sourced from the \emph{YouVersion} online platform\footnote{\url{https://www.bible.com/en-GB/}.}. The Bibles dataset consists of 188,033 parallel verses from the New Testament in English and the 39~Indigenous languages listed in~table~\ref{tab:39bils}. The parallelism among translations of the same verse, performed by some of the authors, is of reasonable quality, although we are aware that the sources of the translations come from different versions, languages, and narrative styles of the Bible.

To avoid cross-contamination in the decoder and to study memorization issues without data leakage between the training and test sets, we used the \emph{Matthew} chapter from the New Testament as the source for the test set, and the remainder of the text as the training set. 
%We used all the books of the New Testament (about 2,800 sentence-pairs) plus Old Testament books for training, and the whole book of Matthew (970 sentence-pairs) for testing. 
%FROMREBUTTAL
%The choice of using a single book was made to avoid that any error in the alignment of the Indigenous and the English versions to be propagated into the test set. 
We are aware that some similarity among verses could happen between the book of Matthew and the other \textit{synoptic gospels}, such as \textit{Mark} and \textit{Luke}. However, we consider the existence of some similarity to be positive, as in most practical multi-language training datasets the presence of similar sentences is common. %Also, the size of book was deemed to be good for a test set since it is one of the longest New Testament book, with 28 chapters.

We understand that using the Bible as a source of data for Indigenous languages raises important ethical, historical, and cultural concerns.
Many unfortunate aspects of past and present colonial history of Indigenous peoples, especially in the Americas,  are connected to different forms of Christianity. At the same time, the Bible is one of the most commonly found documents translated to several of those languages, by Jesuits in the early days of colonization and often by Evangelical churches and initiatives over the last 100 years~\cite{franchetto2008guerra}. As such, the translations of the Bible are often associated with different forms of cultural abuse and violence and the establishment of \textit{orthographies of domination}~\cite{franchetto2008guerra}. 

However, such texts are one of the few available sources of parallel multilingual datasets for most Indigenous languages. We thus view the use of the Bible in this work as an ``exceptional'' first step, where it is treated as potentially ``toxic'' data that should not be used, in principle, for any actually deployed system unless with explicit agreement of the Indigenous community. Nevertheless, we believe Bible data can be used carefully for in-laboratory technical experiments in well-contained contexts, such as the studies we have performed. To mitigate some of these risks, we implemented the protocols suggested in~\cite{pinhanez2023balancing}, including the adoption of containment procedures.

\subsection{The Perils of Multilingual Translators}

To evaluate whether using multilingual MTs was a effective strategy for endangered Indigenous languages, as often suggested by the literature~\cite{lee-etal-2022-pre,chen2022}, we used the \emph{Bibles} dataset, and fine-tuned two commonly-used LLMs using one bilingual and two multilingual fine-tuning strategies.

The first pre-trained model is {\bf mBART50}~\cite{tang-etal-2020-mbart50},  an extended version of \emph{mBART}~\cite{liu-etal-2020-mbart}, with 680M parameters, and pre-trained with masked language modeling on 203M sentences. %This model is a common choice for training multilingual MTs for low-resource languages.
The second LLM is {\bf WMT19}~\cite{ng-etal-2019-facebook}, a 315M-parameter German-to-English machine translator pre-trained on about 28M pairs of translated sentences and over 500M back-translated sentences. For more details about the experimental methodology, refer to~\cite{cavalinnaacl24}.
%, and it is a commThe results are evaluated with three standard reference-based metrics for machine translation, i.e. the \emph{n-gram} based BLEU score and the trained neural metrics named BLEURT and BERTScore.

Our evaluation considered three different fine-tuning strategies using data from the Bibles dataset, resulting in three different types of models, evaluated on two sets of test data from the same dataset. Firstly, we assessed {\bf bilingual (BL)} models created by fine-tuning each LLMs exclusively on source-to-target pairs from  the BILs listed in table~\ref{tab:39bils}, yielding 39~unique bilingual models. Secondly, we considered the extreme multilingual scenario by fine-tuning both LLMs with {\bf all languages (AL)} at once. Lastly, we created in-between multilingual solutions, the {\bf Tupi-family (TF)} models, where the training set comprised by 10~languages belonging to the \textit{Tupi-Guarani} family.

\begin{figure}[t!]
    \centering

    \includegraphics[width=7.6cm]{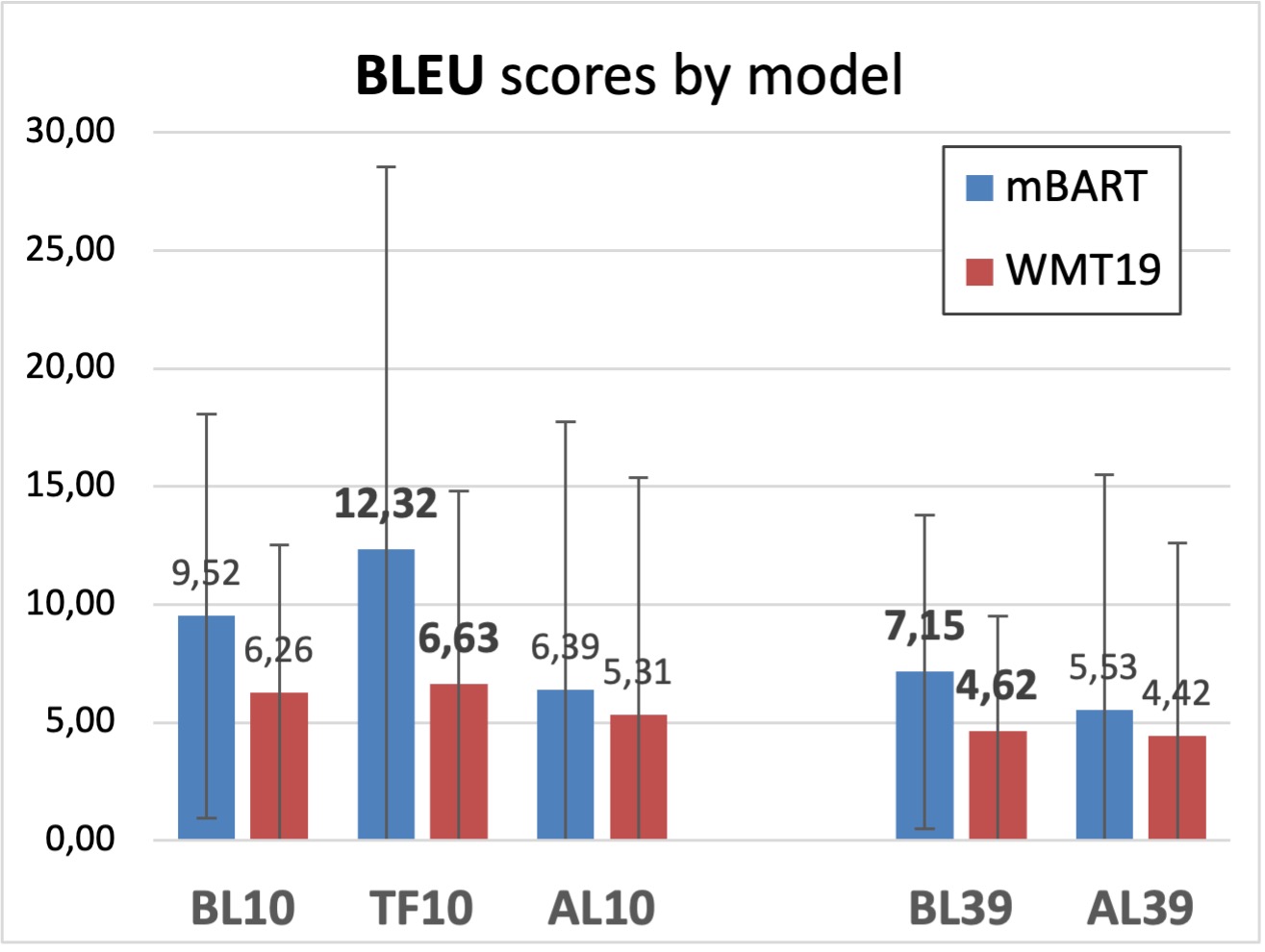} 
    \\ \vspace{2mm}
    \includegraphics[width=7.6cm]{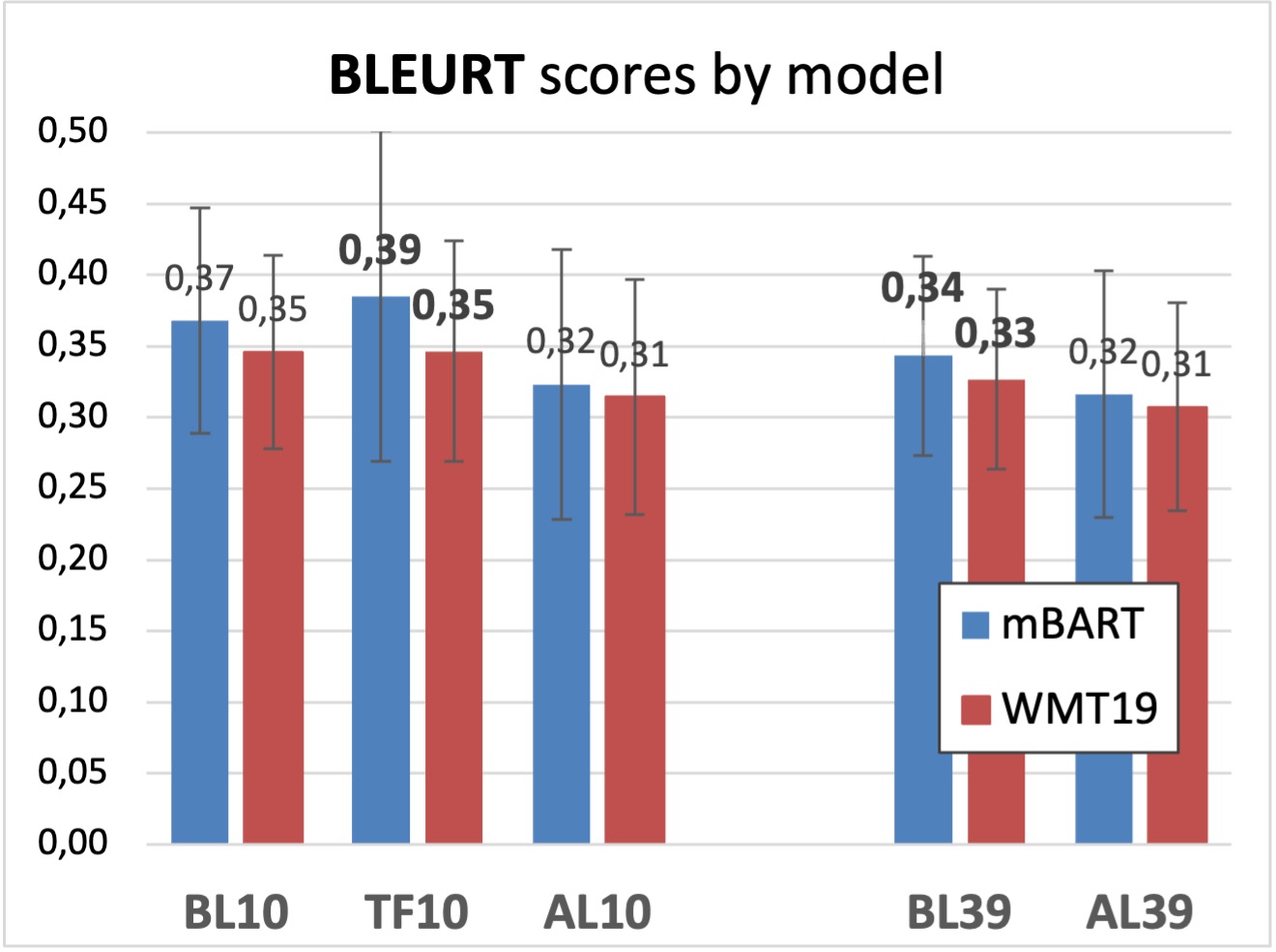} 
    \\ \vspace{2mm}
    \includegraphics[width=7.6cm]{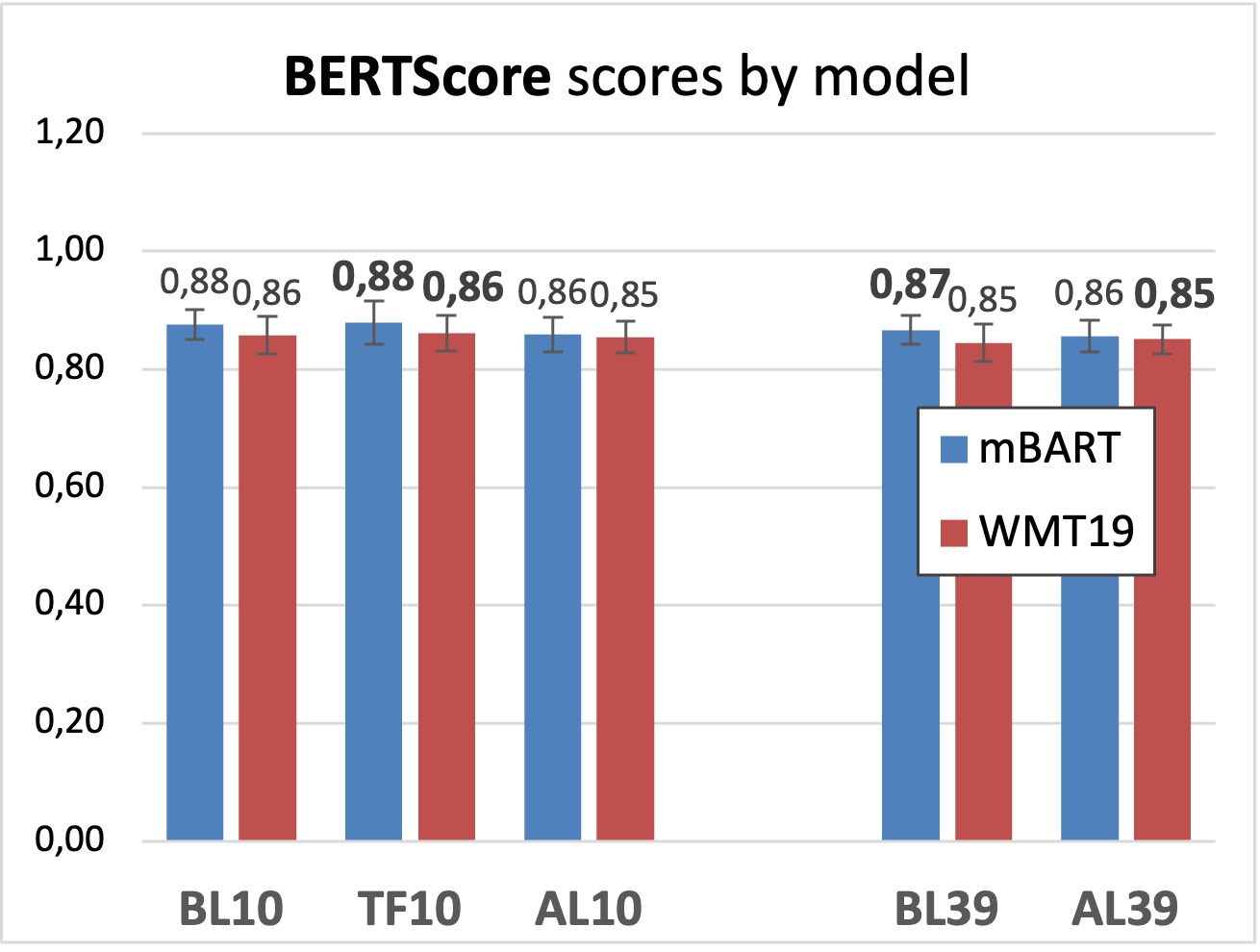} 
    \\  \vspace{2mm}

    \caption{Average and standard deviation scores in different test sets of languages for the fine-tuning of mBART50 and WMT19 to different models for each of the three metrics. The best models of each group are indicated with the bold typeface.
    }
    \label{fig:translators_socores}
\end{figure}

With the goal of measuring the impact of the previously-mentioned models, we defined two distinct sets of experiments. The first one, {\bf BL39 vs. AL39}, considered all 39 BILs in a single test set, so that we could compare the performance of each bilingual model on its own language test set against a multilingual one trained with all the languages for each language data set. The second set is {\bf BL10 vs TF10 vs AL10}, where we compared the BL models not only against AL but also against TF, which is more targeted at Tupi-family languages. For this, the test dataset contained only the 10~languages used to train the TF models.

We used three metrics to evaluate the results. %, combining the traditional {\bf BLEU} score~\cite{papineni-etal-2002-bleu} with more recent neural-based metrics which are considered to be more robust and better correlated with human scores \cite{freitag-etal-2022-results}. 
The first is the traditional {\bf BLEU} score~\cite{papineni-etal-2002-bleu}, using the average of sentence-level BLEU scores. This allowed us to compare all metrics with the same methodology, used by two neural-based metrics, {\bf BLEURT}~\cite{sellam2020bleurt} and {\bf BERTScore}~\cite{bert-score}. 

A summary of the results is provided in figure~\ref{fig:translators_socores}.
As it can be seen, mBART50 performed slightly better than WMT19 but with higher standard deviation, and the three metrics afforded similar results. %So, for the sake of simplicity, we will report results only with mBART50 and SacreBLEU hereafter.
%Considering the two strategies to split the training and test sets, 
The three models yielded similar results, but the bilingual ones presented smaller standard deviation in comparison to the AL models. The TF10 model performed slightly better than the bilingual models but with a higher standard deviation, although the differences were not statistically significant.

\begin{figure}
    \centering
    \begin{subfigure}{0.237\textwidth}
        \includegraphics[width=\linewidth]{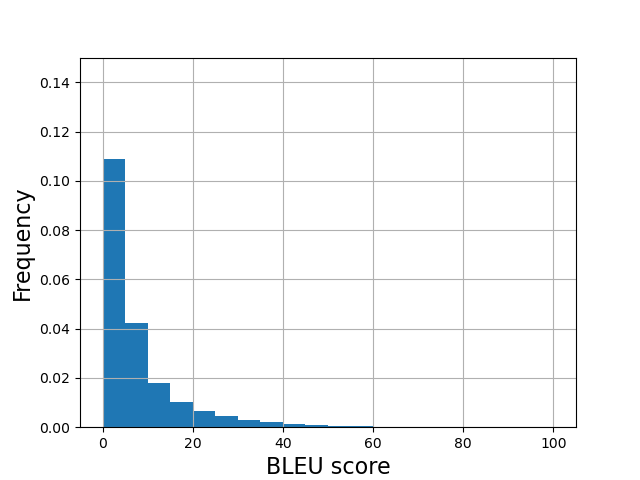}       
        \caption{mBART-BL39}
    \end{subfigure}
    \hfill
    \begin{subfigure}{0.237\textwidth}
        \includegraphics[width=\linewidth]{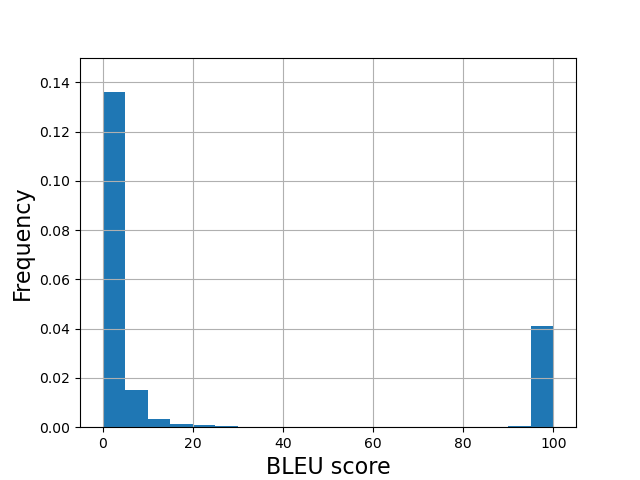}
        \caption{mBART-AL39}
    \end{subfigure} \\
 
    \begin{subfigure}{0.237\textwidth}
        \includegraphics[width=\linewidth]{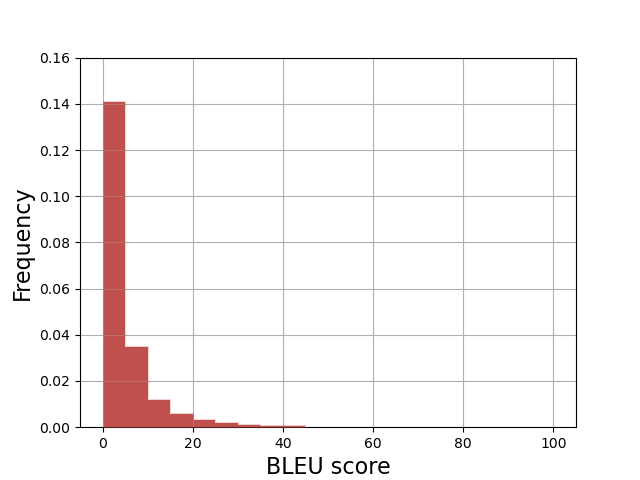}       
        \caption{WMT19-BL39}
    \end{subfigure}
    \hfill
    \begin{subfigure}{0.237\textwidth}
        \includegraphics[width=\linewidth]{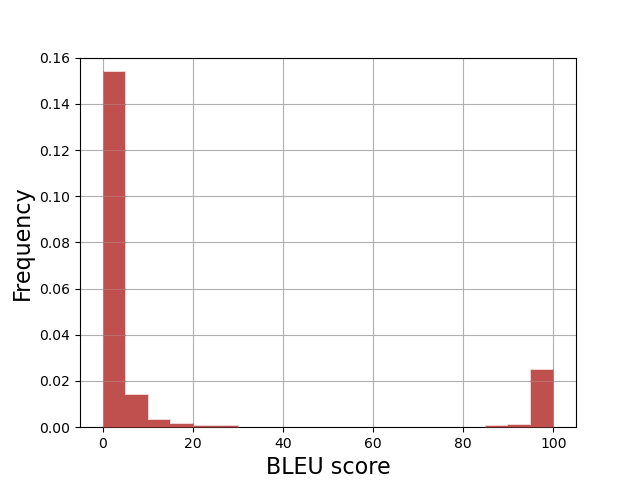}
        \caption{WMT19-AL39}
    \end{subfigure}

    \caption{Distribution of BLEU scores of samples from the training set for mBART50 (top) and WMT19 (bottom). The peaks on the right of the AL39 histograms, corresponding to perfect scores, are evidence of rogue memorization strategies}
    \label{fig:bleu_distributions}
\end{figure}

In this work, multilingual MTs did not seem to perform better than bilingual translators. Therefore, our conclusion was that in cases where there is just the need for a translator for one language or data is available for only one language, working with a bilingual translator seems to be the natural choice. However, it could be argued that, in cases where data from multiple languages is available, the converse, that is, working with multilingual translators, would be recommended. As described in~\cite{cavalinnaacl24}, this is not the case, where we showed that high scores of the multilingual translators were achieved not by improving the quality of the translation, but by learning a ``wrong'' strategy of memorization and retrieval, signaled by the higher standard deviations.

In fact, we found that high standard deviations were a symptom that the translation model had started to perform \emph{rogue memorization}, that is, it had become a retriever of the contents of the training set instead of a translator~\cite{cavalinnaacl24}. This can be seen by looking at the distribution of BLEU scores computed with samples from the training set, shown in figure~\ref{fig:bleu_distributions}. From these distributions, we can clearly observe that the BL39  models (both for mBART50 and WMT19) usually result in right-skewed normal distributions, while for AL39 the shapes of the distributions resemble more binomial distributions.

The high number of training samples with perfect translations of those models on the right side of the histograms were obtained by the models adopting a erroneous strategy of storing internally some of the verses, as further demonstrated in~\cite{cavalinnaacl24}. The paper also shows that the cause of this pernicious learning behavior was connected to the many-to-one mapping of the multilingual training datasets which can be alleviated by the use of rephrasing.

\subsection{The Impact of Data Quantity}

%SIGUL 24 paper

While working with the Tenondé-Porã community and their commitment to enabling their high-school students to write in Guarani Mbya, we considered that having translators to and from Portuguese would be helpful. 
%. The \emph{Guarani Mbya} language is a language of the \textit{Tupi} family, spoken by approximately 8,000 people, mostly in the South-Southeast area of Brazil, and, although being a language still actively spoken and well-studied, it has very few sources of translated texts which can be used to mine bilingual pairs of sentences essential for the training of today's ML translators. Guarani Mbya is similar to the \textit{Guarani} language spoken by about 1~million people in Paraguay, but too different to allow the use of data from that language.
Following our experiences with fine-tuning MTs in the context of the Bibles dataset, we started developing a Guarani Mbya to English MT to be used in conjunction with a commercial English-Portuguese translator. 

To train the MT, we created a dataset, referred to as the \emph{Dictionary} dataset, with pairs of sentences from three different sources. The first source was a set of Guarani Mbya short stories with 1,022 sentences, available in Portuguese and English~\cite{dooley1988a, dooley1988b}. The second source comprised 245 texts extracted from PDF files with a pedagogical character \cite{dooley1985}. The third source was Robert A. Dooley's \emph{Lexical Guarani Mbya dictionary}~\cite{dooley2016}, a reference work for the language, from which we extracted 2,230 sentence pairs.
%, and the reason why the dataset was named Dictionary. %The last two sources contained sentence pairs in Guarani Mbya and Portuguese only. We converted them to English using a Portuguese-to-English commercial translation service. We have permission from the author to use this data.
%After concatenating the data from the three sources, we cleaned it, removing some non-alphanumeric characters (e.g. *, $\gg$, •) and normalizing Unicode values. Then, the Dictionary dataset was split 
In total, the Dictionary dataset had 3,155 training and 300 test sentence pairs.

We then considered using the Guarani Mbya subset of the Bibles' dataset to increase the amount of training data. Keenly aware of the ethical issues of using texts from the Bible in a translator for Indigenous languages in Brazil, we conducted a series of experiments,  detailed in~\cite{dominguessigul24}, to determine whether the additional ``toxic'' data from the Bible was useful and, if so, how much its use would contaminate the outputs.

As a baseline for this study, we defined the \texttt{zeroshot} model, consisting of the original German-English \texttt{WMT19} model~\cite{ng-etal-2019-facebook} without any fine-tuning. Using only the Bibles training set, we generated three different models based on directly fine-tuning \texttt{WMT19}: \texttt{mbya}, the \texttt{WMT19} model fine-tuned with only the Guarani Mbya data from the Bibles training set; \texttt{TGf}, the \texttt{WMT19} model fine-tuned with Bibles data from 10~languages of the \emph{Tupi-Guarani} linguistic family; %\emph(\emph{Guarani} of Paraguay and Bolivia (2);  \emph{Guarani Kayow\'a}, \emph{Guarani Mbya}; \emph{Ka'apor}, \emph{Kaiabi}, \emph{Nheengatu}, \emph{Guajaj\'ara}, and \emph{Tenharim}, aiming to take advantage of the geo-linguistically similarity of those languages;
and \texttt{all}, the \texttt{WMT19} model fine-tuned with data from all the 39~Indigenous languages of the Bibles training set. %(E5)

\begin{figure*}[t!]
    \centering

    %\includegraphics[width=7.6cm]{figures/mbya_dict_test_sacrebleu.jpg} \\ 
    %\vspace{2mm}
    \includegraphics[width=7.6cm]{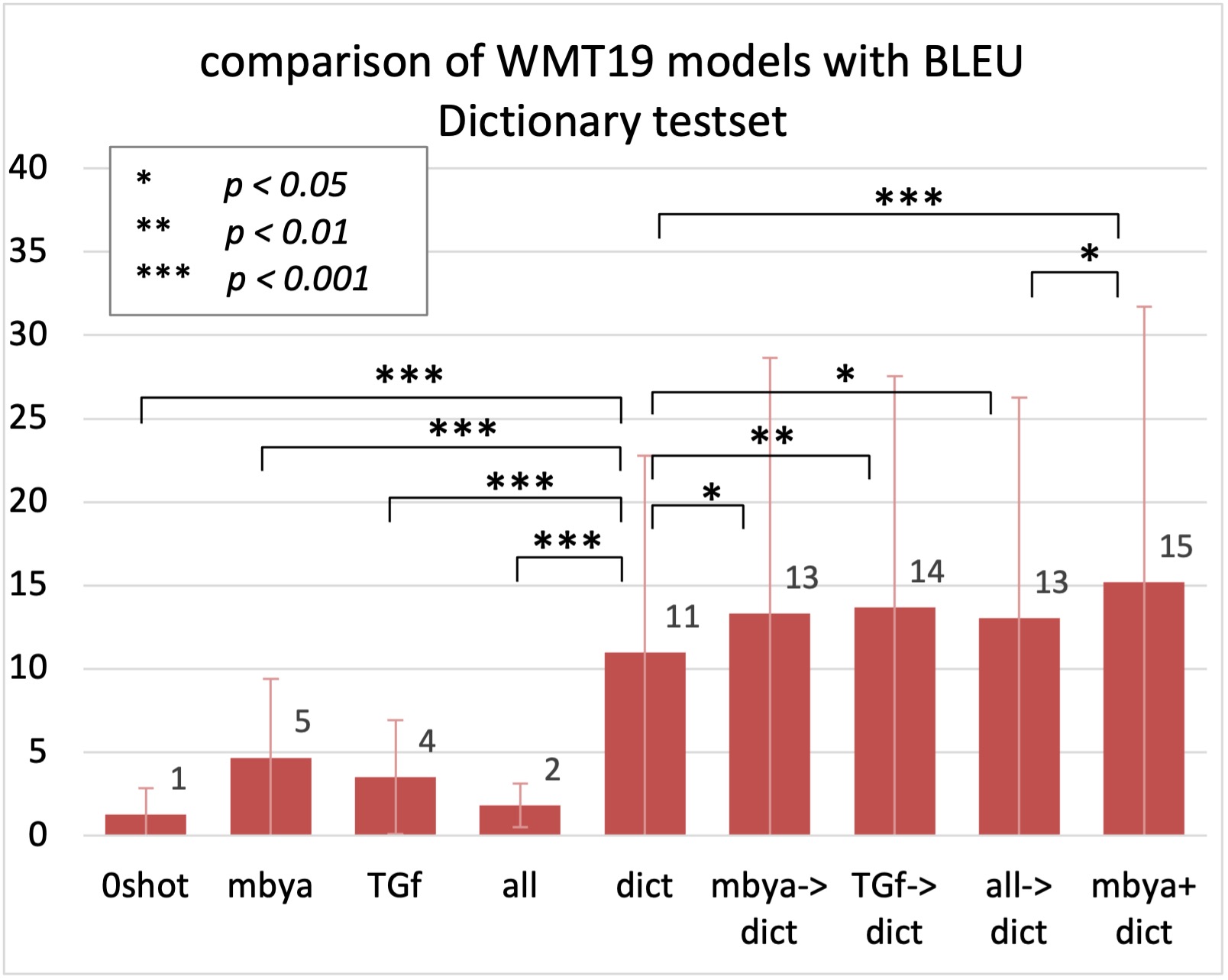} 
    %\\ \vspace{2mm}
    \hspace{2mm}
    \includegraphics[width=7.6cm]{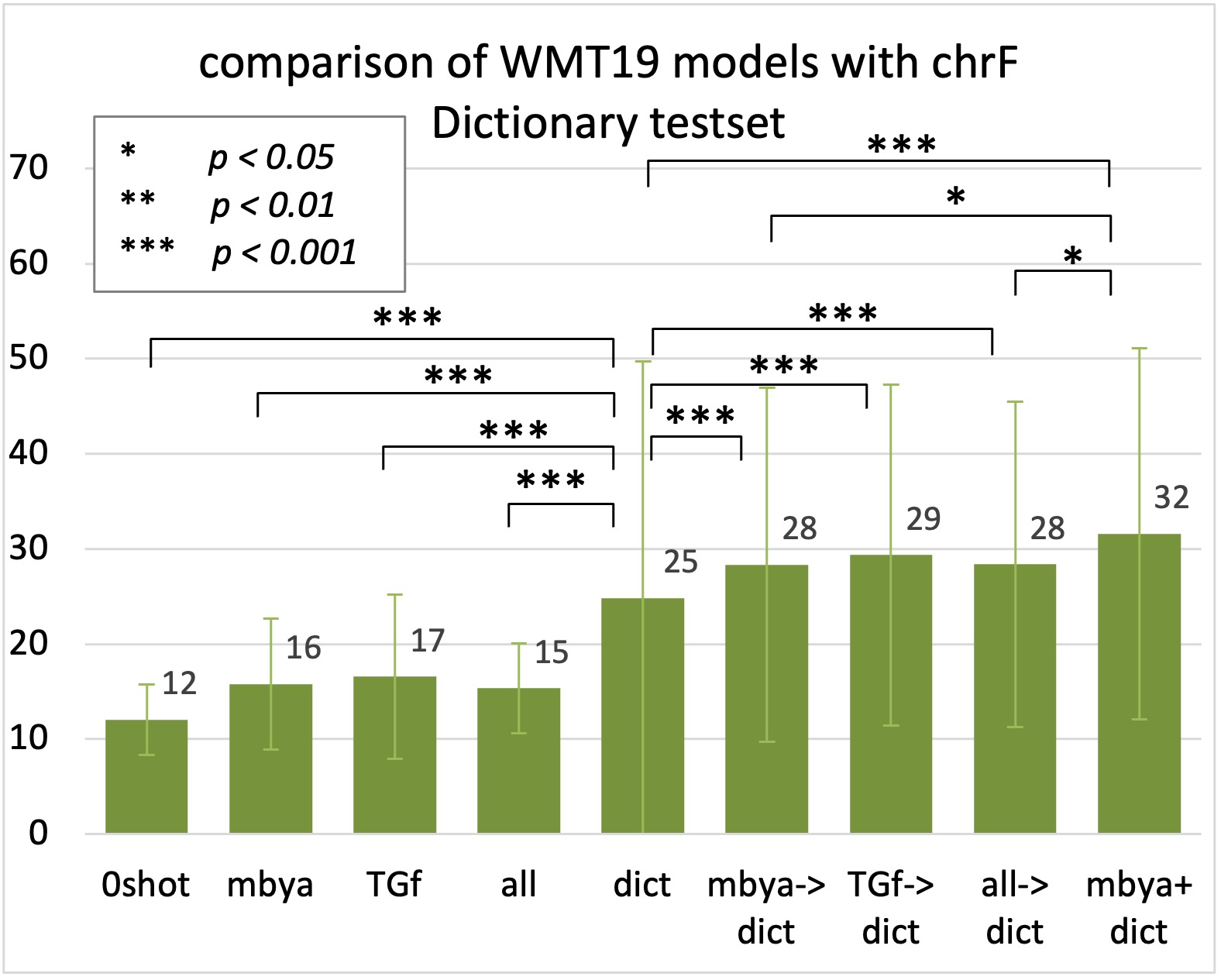}

    \caption{Performance according to the metrics SacreBLEU (top) and chrF (bottom) in the Dictionary test set of the original WMT19 model and its fine-tuning into 8~Guarani Mbya translators using different training data sets and regimes; significant differences are also shown.
    }
    \label{fig:mbya_dict_tests}
\end{figure*}

Using the data from the Dictionary training set, we generated four additional models: \texttt{dict}, the \texttt{WMT19} model fine-tuned only with Dictionary data; \texttt{mbya>dict}, the \texttt{mbya} model fine-tuned a second time with Dictionary data; \texttt{TGf>dict}: the \texttt{TGf} model fine-tuned a second time with Dictionary data; and \texttt{all>dict}: the \texttt{all} model fine-tuned a second time with Dictionary data. Finally, we trained \texttt{mbya+dict}, consisting of the \texttt{WMT19} model fine-tuned with Guarani Mbya data from the Bibles training set and Dictionary data simultaneously.

We used two metrics to evaluate the results: %{\bf BLEU} score~\cite{papineni-etal-2002-bleu}, computed with the NLTK package\footnote{https://www.nltk.org/}, with default parameters and metric-defined tokenization;  
the {\bf BLEU} metric, computed with the SacreBLEU Python package~\cite{post-2018-call};  and the {\bf chrF} metric~\cite{popovic-2015-chrf}, in both cases using segment-level scores. % which, although being a metric for poly-synthetic languages, has been widely applied in recent works with low-resource languages. 
For the two metrics, we computed the average and standard deviation over the score of each sentence in the two test sets created from the Dictionary and Bibles datasets.

Figure~\ref{fig:mbya_dict_tests} shows the results of the 9~models, evaluated with the Dictionary test data. For the two~metrics, the \texttt{zeroshot} model had an extremely low performance, as expected since it is basically a German-to-English translator. The performance of the three models fine-tuned with Bibles data was poor, as expected. This becomes clearer when comparing the \texttt{dict} model to them:
%and to the original \texttt{zeroshot}: 
the average accuracy was considerably improved. Although \texttt{dict} had a larger standard deviation, it was significantly better than the other models ($p < 0.001$) for the 2~metrics, using standard \emph{one-tailed Student t-tests}.

The three two-step models (marked with \texttt{->dict}), had gains of about 16\% to 36\% in accuracy over \texttt{dict}. The t-tests confirm that each of these models was significantly better than \texttt{dict}. The best nominal performance was achieved with the both-at-once model, \texttt{mbya+dict}, across all metrics and test sets, although there was some statistically significant difference ($p < 0.05$) compared to some of two-step models.

The results with the Dictionary test set indicated, with high confidence and across all metrics, that the best results were achieved by the fine-tuning of the \texttt{WMT19} model with the two types of data, from linguistic resources and the Bible, at the same time. In other words, the quantity of data available for fine-tuning seems to matter, as expected, and the best training methodology involved using  all the data together in a single fine-tuning process. A more detailed discussion and related results are available in~\cite{dominguessigul24}.

\subsection{Quantifying Contamination}

Having established that adding the Bible data could improve the accuracy of the Guarani Mbya to English translator, even when tested only with non-Bible sentences, the work then focused on quantifying the extent of contamination in the output. Specifically, we aimed to determine how many of the translators' outputs contained, either explicitly or implicitly, typical words or language from the Bible.

\begin{table*}[t!]
\centering
\includegraphics[width=\textwidth]{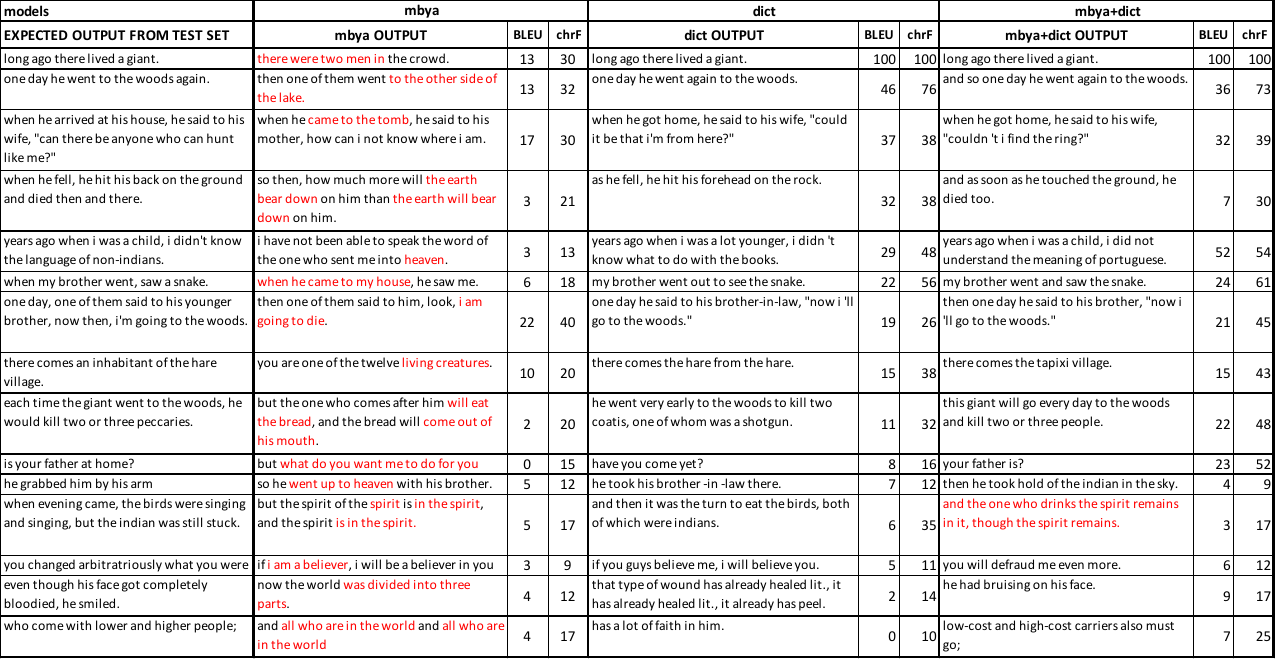}
\caption{Examples of outputs of the mbya, dict, and mbya+dict models with SacreBLEU (marked as BLEU) and chrF scores and the expected output from the test set; segments which are associated with biblical texts and expressions are marked in red.}
\label{tab:outputs_dict_mbya+dict}
\end{table*}

\begin{table*}[t!]
\centering
\includegraphics[width=\textwidth]{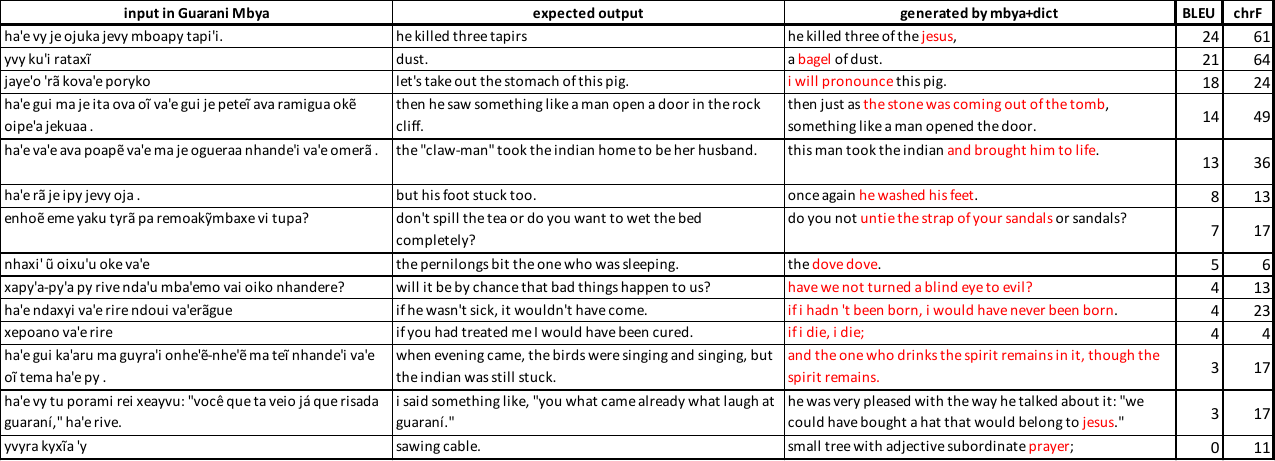}
\caption{Outputs of the mbya+dict model which were identified as possible cases of contamination; segments which are possibly associated with biblical texts and expressions are marked in red.}
\label{tab:contaminated_outputs}
\end{table*}

Table~\ref{tab:outputs_dict_mbya+dict}, from~\cite{dominguessigul24}, contains a random selection of outputs for 15~sentences in Guarani Mbya, together with the expected output. Segments manually identified as associated with biblical texts and expressions are marked in red, for three of the models. All the 15~outputs of the \texttt{mbya} translator have contamination, as expected since it was trained only with Bible data, while the \texttt{dict} translator, which was not exposed to Bible data in the fine-tuning process, had none.

The \texttt{mbya+dict} outputs depicted in table~\ref{tab:outputs_dict_mbya+dict} show only one possible case of contamination.  This output,  \emph{``and the one who drinks the spirit remains in it, though the spirit remains.''}  resembles the \textit{John 6:56} verse, \emph{``The one who eats my flesh and drinks my blood resides in me, and I in him.''}.

The qualitative evaluation of all the 300~outputs of the \texttt{mbya+dict} for the Dictionary test set found that 14~(4.7\%) of the~300 outputs had some level of contamination, including 2~obvious cases where the word \emph{``Jesus''} appeared. These 14~outputs are shown in table~\ref{tab:contaminated_outputs}, from~\cite{dominguessigul24}. They include, as contaminated outputs, examples where words such as \emph{``prayer''}, \emph{``dove''}, \emph{``bagel''} were produced; and expressions such as \emph{``washed his feet''}, \emph{``blind eye to evil''}, and \emph{``if I die, I die''}, whose degree of connection to the Bible is questionable.

Based on those findings, as discussed in \cite{dominguessigul24}, we advise against the release of the \texttt{mbya+dict} translator in broader contexts and recommend its use only in tightly controlled situations where negative effects can be mitigated. Of course, the final decision shall be made by the relevant Indigenous communities~\cite{mihesuah1993suggested,sahota2007research,straits2012guiding}. 

\subsection{The Impact of Data Quality}

In our collaboration with the high school students of the Tenondé-Porã community, we incorporated the \texttt{dict} Guarani Mbya to English translator with a commercial, API-based English to Portuguese translator into a writing assistant, which will be discussed in detail in section~\ref{sec:writing-assistant}. As expected, due to the low SacreBLEU and chrF average scores (see figure~\ref{fig:mbya_dict_tests}), the quality of the translations it generated were of limited usefulness to the students.

Improving the quality of this translator was important, so we began by seeking a better understanding of the errors it was producing and assessing the usefulness of different parts of the output. Drawing conclusions about human usefulness of a translator based only on values from automatic metrics such as SacreBLEU is challenging, since they rely on straightforward computations such as word comparison and $n$-grams, often overlooking semantic issues. Therefore, to determine the usefulness of the translators, we conducted a human evaluation on the texts generated from the test set inputs. 

This evaluation, described in detail in~\cite{pinhanezillc24} involved the ranking of each of the generated outputs in a seven-point scale according to how useful it would be for someone proficient in the language using the translator as a writing support: \textit{near-perfect}, \textit{correct}, \textit{mostly correct}, \textit{usable}, \textit{mostly incorrect}, \textit{incorrect},  and \textit{very wrong}.

\begin{figure*}[t!]
    \centering

    \includegraphics[width=7.0cm]{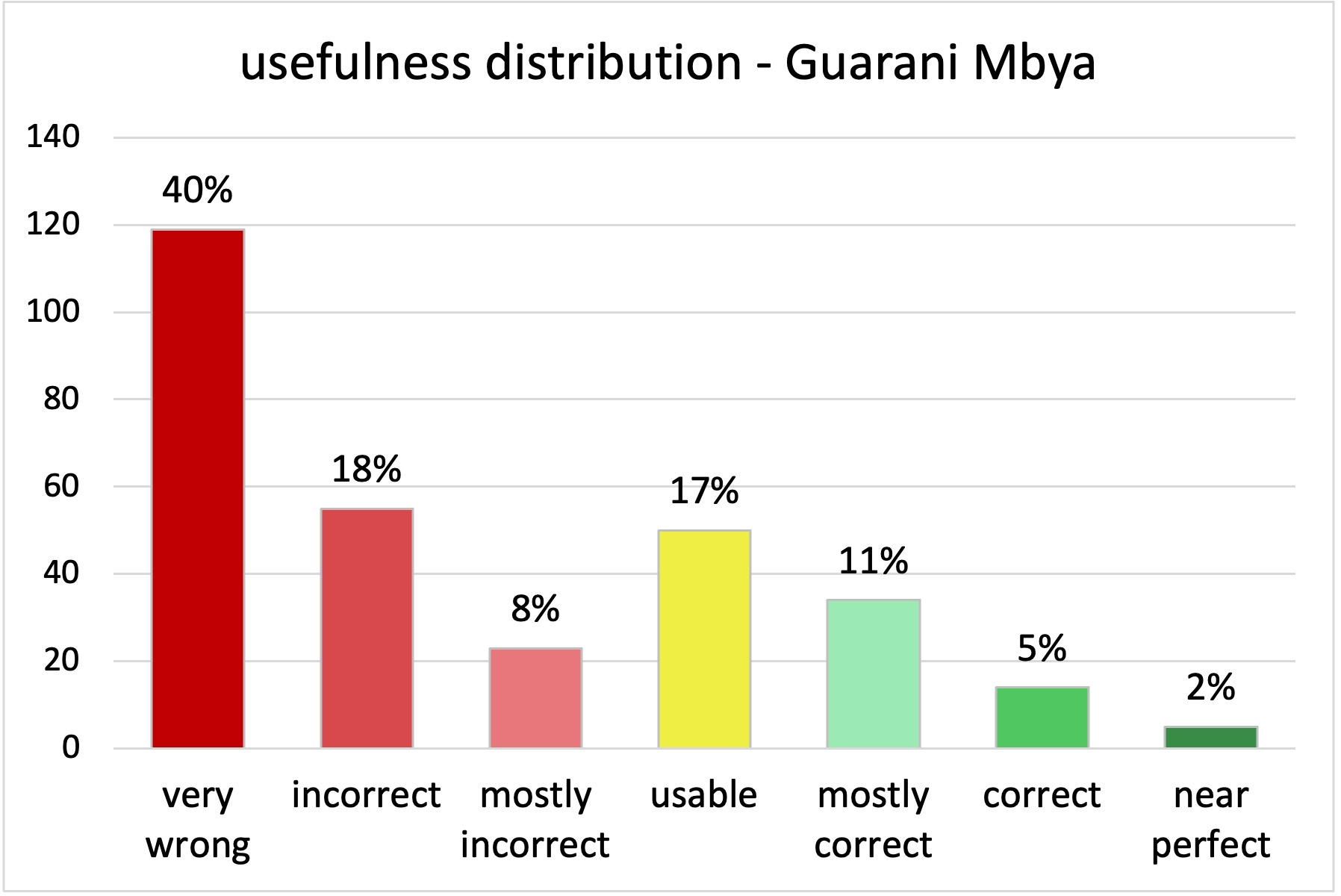} 
    \hspace{2mm}
    \includegraphics[width=7.6cm]{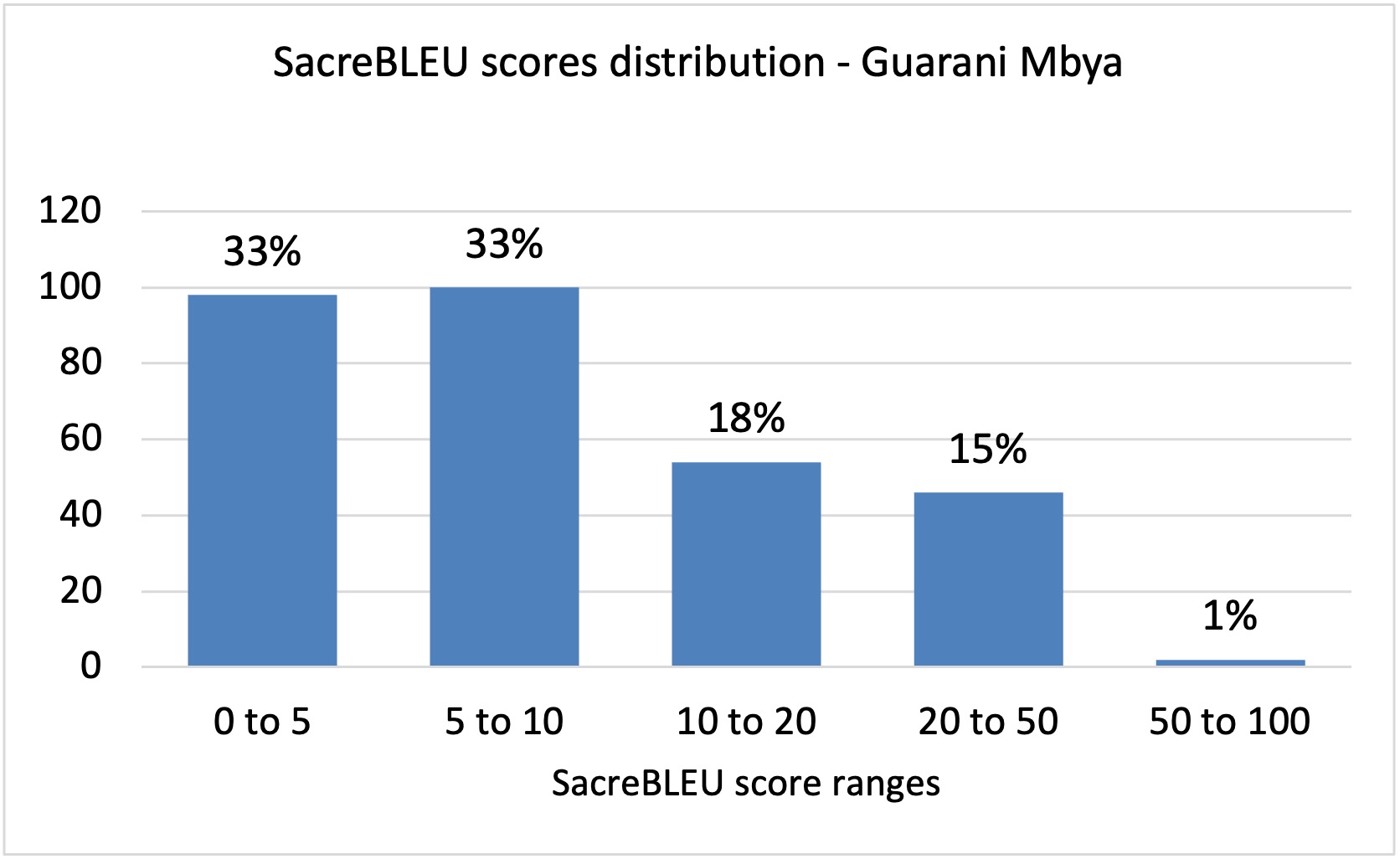}

    \caption{Histograms of the performance of the Guarani Mbya translator according to the human evaluation of usefulness of the outputs (left) and, as a reference, the SacreBLEU metrics (right).
    }
    \label{fig:mbya_usefulness}
\end{figure*}

\begin{figure*}[t!]
    \centering

    \includegraphics[width=7.1cm]{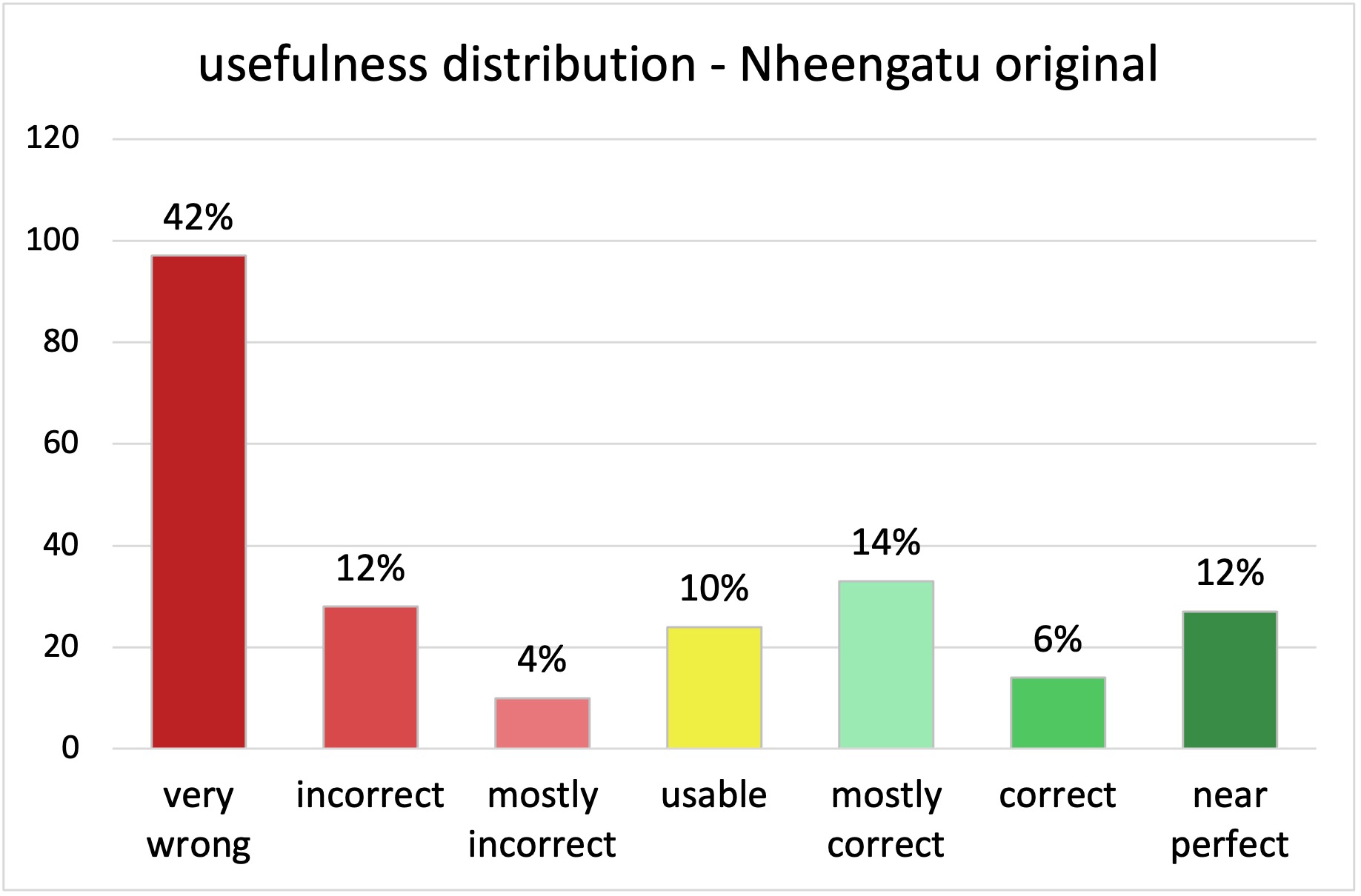} 
    \hspace{2mm}
    \includegraphics[width=7.6cm]{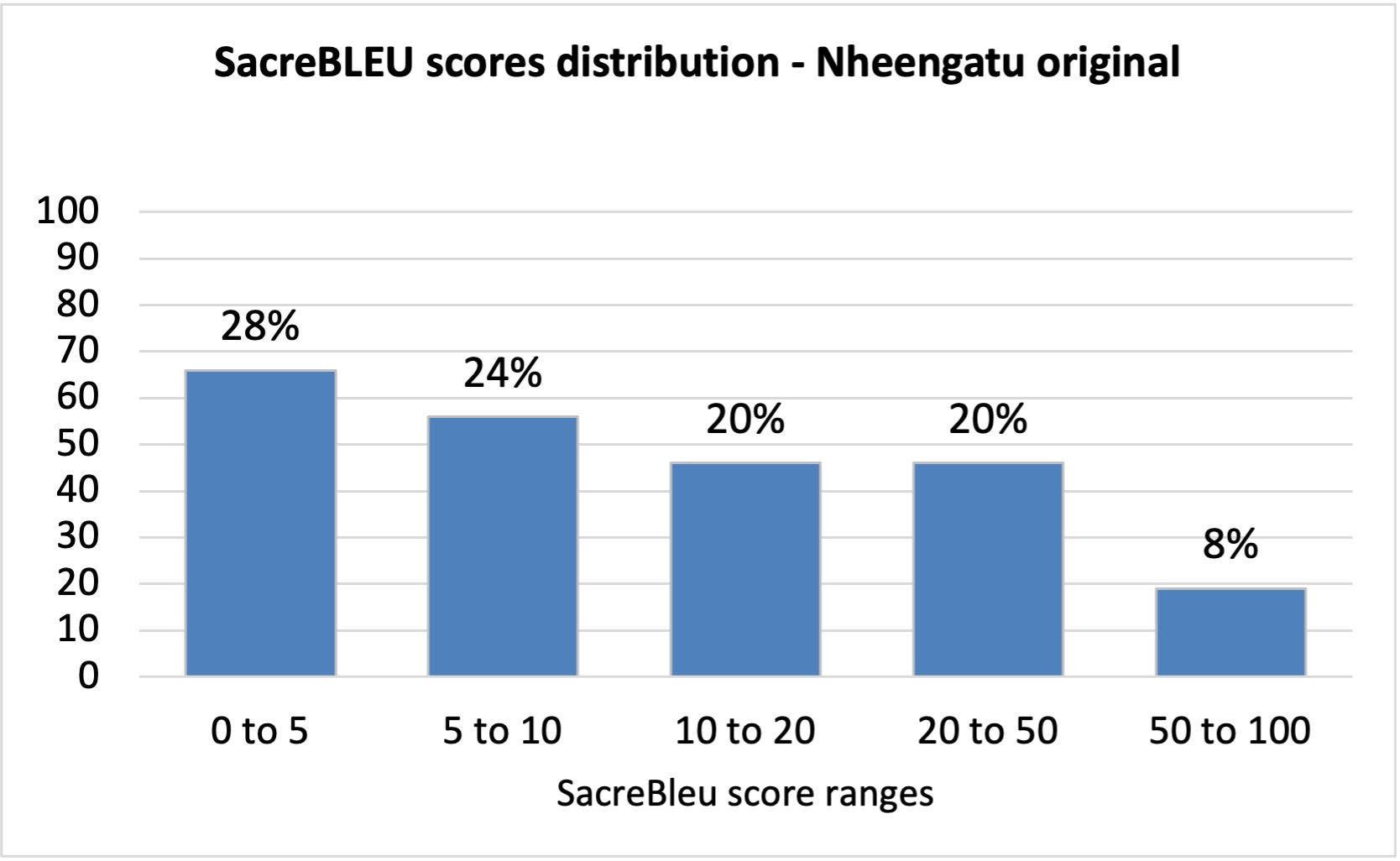} 

    \vspace{2mm}

    \includegraphics[width=7.1cm]{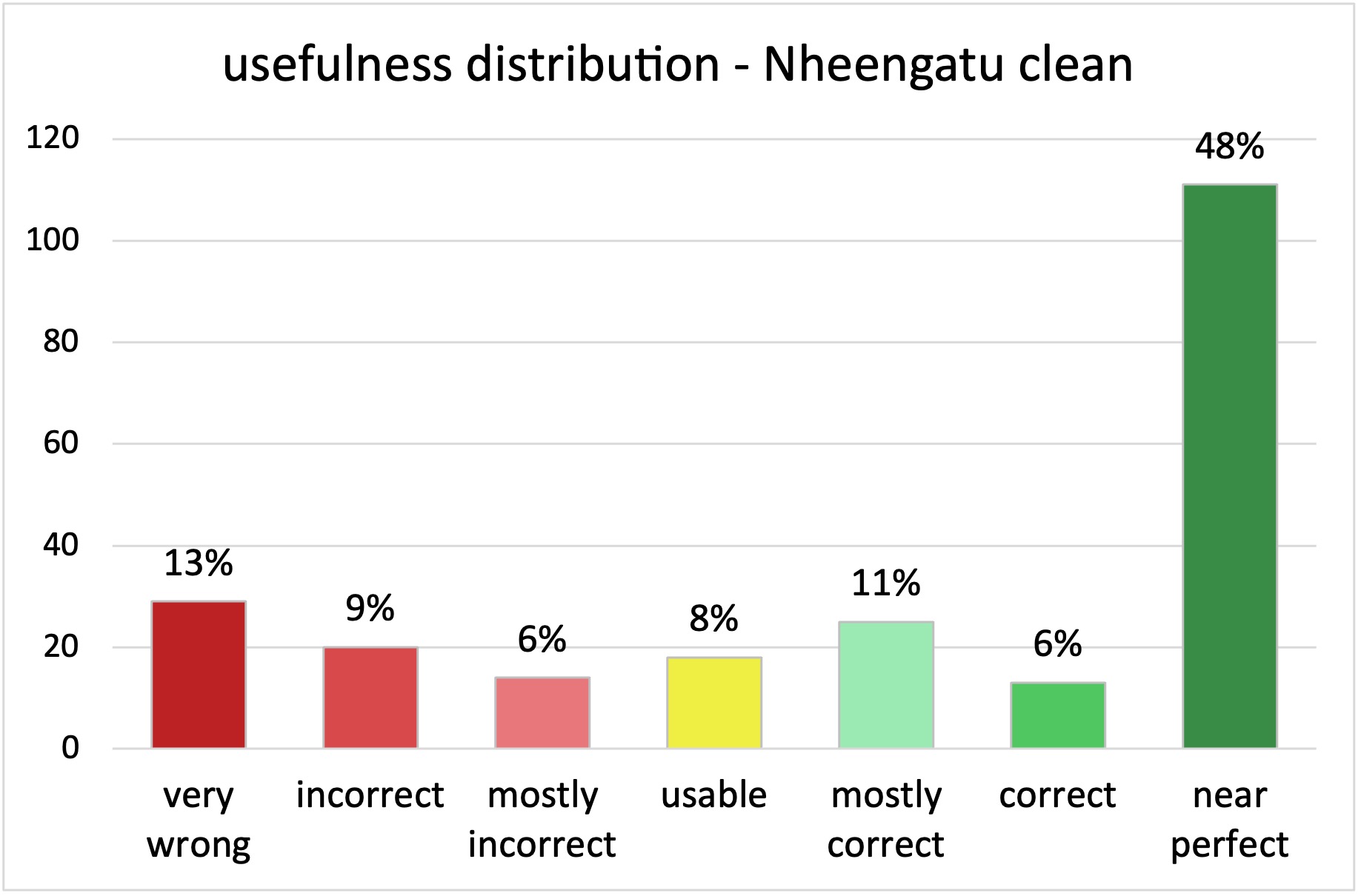} 
    \hspace{2mm}
    \includegraphics[width=7.6cm]{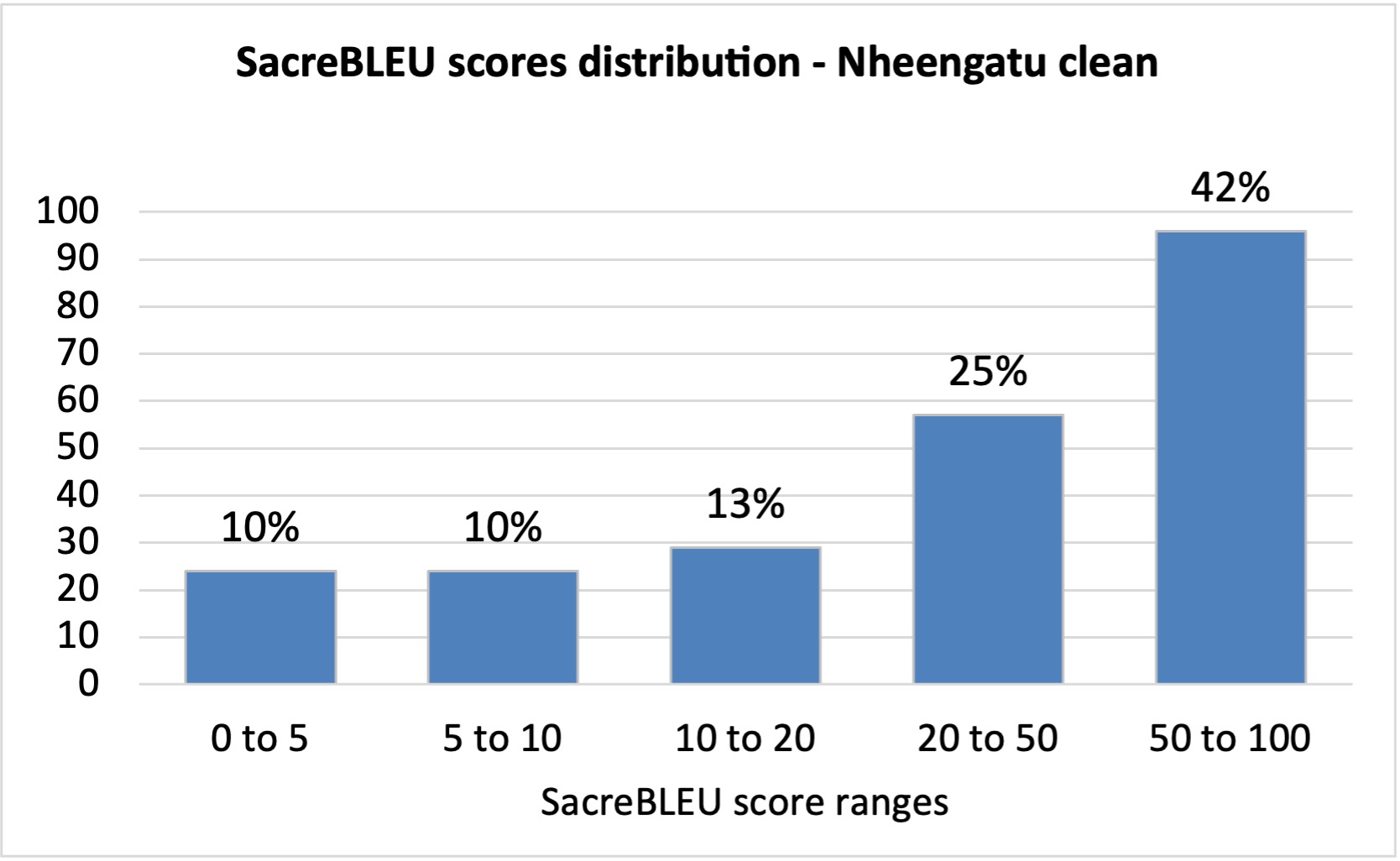}

    \caption{Histograms of the performance of the Nheengatu translators trained with the original data (top) and the cleaned  data (bottom) according to the SacreBLEU metrics (left) and the human evaluation of usefulness of the outputs (right).}
    \label{fig:nheengatu_usefulness}
\end{figure*}

Figure~\ref{fig:mbya_usefulness} (left) shows the histogram of the distribution of the 300~outputs of the Guarani Mbya translator (\texttt{dict}), according to the usefulness scale. About 40\% of all outputs were in the \emph{very wrong} category, with 26\% categorized as \emph{incorrect} and \emph{mostly incorrect} categories. Of the remaining 34\%, about 28\% were sentences requiring a significant level of human intervention to be used (\emph{usable} and \emph{mostly correct} categories), with only 7\% suitable for an automatic translation scenario. As a reference, the distribution of the SacreBLEU scores of these 300~outputs is shown in figure~\ref{fig:mbya_usefulness} (right).

As we moved to work with the Nheengatu community, as discussed earlier, we decided to apply the same techniques to develop a Nheengatu to English translator. The \emph{Nheengatu} dataset comprised sentences from five different sources with Portuguese translations. 
The first source was the \emph{Nheengatu lexicon}~\cite{avila2021} with 6,846 sentences extracted from the lexicon examples. We processed the original file provided  by the author. The second source was \emph{Corpus Lições}~\cite{avila2021}, containing 1,665 samples available in a spreadsheet format. The other sources, extracted from PDFs, included: \emph{Texto Anônimo}~\cite{navarro2011}, with 427 samples; \emph{Brilhos na Floresta}~\cite{ishikawa2019}, with 590 samples; and \emph{Curso LGA}~\cite{navarro2016}, with a partial extract of 590 samples.

In total, the Nheengatu dataset contained 7,281 samples, with a random split of 6,804 samples for training and 233 samples (10\% of the data from all sources except Nheengatu lexicon) for testing. We translated the Portuguese sentences to English, when necessary, using a Portuguese-to-English commercial translation service\footnote{IBM Watson Language Translation v9.0.0.}.

The top row of figure~\ref{fig:nheengatu_usefulness} shows the equivalent results of the Guarani Mbya for the Nheengatu translator. The numbers of the Nheengatu translator were slightly better, with 42\% of the 233 outputs in the \emph{very wrong} category but only 16\% in the \emph{incorrect} and \emph{mostly incorrect} categories. Of the remaining 42\%, 24\% would need human correction to be usable, and 18\% would be suitable for an automatic translation scenario. The distribution of the SacreBLEU scores of the outputs, shown on the right side,  was also better than that of the Guarani Mbya translator, though not significantly better.
The improved results of the Nheengatu translator can be attributed simply to the larger amount of training data.

A side effect of the manual evaluation of usefulness of the outputs was that it became clear to the team that there were many errors in the test set, included expected outcomes which were empty sentences and some residual lexical information in some entries. This was mainly caused by the semi-automatic conversion of the input materials, especially from the lexicon dictionary. %~\cite{avila2021}.
This prompted a manual, exhaustive revision of the whole training set, removing incorrect pairs and fixing others, which resulted in the \textit{Nheengatu Clean} dataset, with 6,848~pairs, about 6\% smaller than the original \textit{Nheengatu} dataset. This dataset was split into 6,621~pairs for training and 227~for testing.

The translator created using the Nheengatu Clean dataset was remarkably better than the one created with the original Nheengatu dataset, as shown in the bottom row of figure~\ref{fig:nheengatu_usefulness}. About 48\% of the 227 outputs were in the \textit{near perfect} category, suitable to automatic translation; 17\% were \textit{correct} or \textit{mostly correct}; 8\% were usable; and only 27\% were unusable, summing up the \textit{very wrong}, \textit{incorrect}, and \textit{mostly incorrect} categories. The average BLEU score jumped from $18.9 \pm 16.8$ with the original dataset to $38.6 \pm 47.1$, flipping the histograms of the SacreBLEU scores, as shown by the right graphs of figure~\ref{fig:nheengatu_usefulness}.

Notice that all the above gains were obtained by removing only 6\% of the entries and fixing basic errors in about 10\% of them. Moreover, the manual inspection of the outputs of the original Nheengatu translator seems to indicate that the translator learned to output errors, that is, to generate empty outputs and to include incorrect elements. What we observed was that a fine-tuning process with very little data seems to see errors as valid outputs and learns to generate them. Fine-tuning LLMs with very small amounts of data seem to require that what is presented to the model to be heavily curated, because the LLMs learn to generate errors if they see even few examples of them.

These and the previous results seem to indicate three important guidelines when building MTs by fine-tuning large translators with very limited amounts of data. First, unlike what is recommended by most of the previous literature, there is almost no gain from using multilingual translators. Second, problematic data sources should not be used since contamination is likely to appear in the outputs. And third, the small datasets used for fine-tuning should be thoroughly inspected and cleaned from errors. %We still have to validate those findings to other languages, including the Guarani Mbya translator generated with the Dictionary dataset.

%%%%%%%%%%%%%%%%%%%%%%%%%%%%%%%%%%%%%%%%%%%%%%%%%%%%%%%%%%%

\section{Using AI to Create Writing Assistants} \label{sec:writing-assistant}

In the second workstream of our project we are developing technologies for the creation of digital \textit{writing assistants} which can simplify and foster the use of  Indigenous languages among their communities, especially teenagers and young adults. This need was identified in our work with both the Guarani Mbya and Nheengatu communities.

Stimulating the use of writing is also, as discussed before, one of the effective methods to diminish the process of forgetting a language learned by individuals as a child. This is true both for languages which were learned orally from family members, friends, and community and also in cases where the formal education process had actually thought them how to read and write. Moreover, in the context of increasing use of digital tools and Internet access, providing easy ways for writing in Indigenous languages in those media is likely to be key to reach teenagers and young adults, often the heaviest users of such tools.

\begin{figure*}[t!]
     \centering
    %  \includegraphics[width= 7.5 cm]{figures/america_languages_histogram.png} 
    %  \hspace{2mm}
    % \includegraphics[width= 7.5 cm]{figures/brazil_languages_histogram.png} 
    \includegraphics[width= 10 cm]{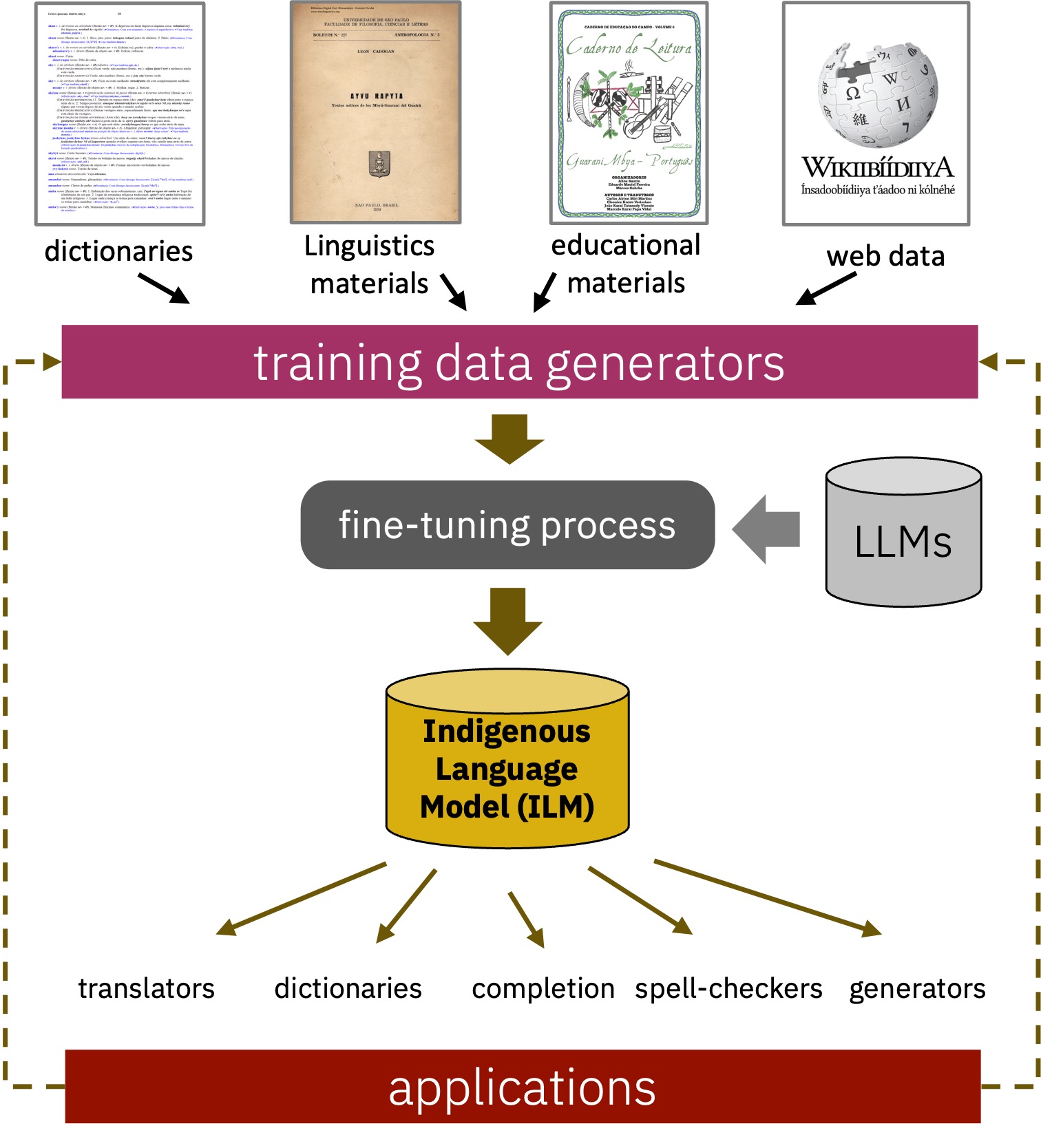} 
     \caption{Technical strategy to develop writing assistants for low-resource Indigenous languages.}
    \label{fig:technical-strategy}
\end{figure*}

We have been developing such writing assistants within the communities using a process of co-design where different \textit{component tools} are prototyped, put together, and tested with Indigenous speakers in the context of different tasks. How those tools are assembled, integrated, and accessed is an essential part of the process, as well as the development of the tools themselves. The main tools we have been building are:
\begin{description}[noitemsep]
    \item[Word dictionary:] a tool to provide access to words, their meanings, and translations based on approximate search;
    \item[Word completion:] a tool which suggests words that can complete a partially-typed word;
    \item[Next-word completion:] a tool which suggests words which can follow a partially-typed sentence; 
    \item[Spell checker:] a tool which suggests corrections in the words of a partially- or a fully-typed sentence;
    \item[Translator:] a tool which translates words and sentences to and from the Indigenous language and another language, or between different orthographies of the same language.
\end{description}

In many ways, those are the typical writing-support tools which languages with a large number of speakers have easily available in word editors, social media apps, and similar. They are often built by dedicated, large professional teams, using traditional programming and machine learning methods, with the support of deployment and maintenance people. We do not believe that such resources are likely to be made available to endangered Indigenous languages, what has prompted us to develop an strategy to build those tools in the context of limited resources and data.

Figure~\ref{fig:technical-strategy} depicts an overview of the technical strategy we have devised based on developing \textit{Indigenous Language Models (ILMs)} by fine-tuning existing LLMs of more resourced languages. The fine-tuning process uses linguistic and educational data as well as data from the Internet, extracted and formatted by a set of \textit{training data generators}, so the ILM is trained to perform the tasks of the component tools described before, such as word completion and spell checking. As the tools are incorporated into applications and used by the community, the new generated data can be used to improve the ILMs and the tools.

There are two structuring premises behind the proposal of this strategy. The first premise is the admission that there are very limited data resources for a given endangered Indigenous language and therefore they should be used efficiently and judiciously. The lack of data clearly pushes towards a fine-tuning strategy instead of a full training model for which there would not be enough data. At the same time, as discussed before, fine-tuning data must be as correct and clean as possible, suggesting as much as possible the use of linguistic data such as dictionaries, lexicons, theses, education materials, and carefully collected web data.

The second key premise, and possibly the most controversial, is the use of an ILM as the basis for all tools. This ILM can be tailored to produce the outcomes of the different component tools by the use appropriate prompting. Using different prompts, the same model can produce translations, provide word and sentence completions, generate answers to questions, or simply retrieve the meanings of words. 

The main reason to adopt this technical strategy  is \textbf{replicability}: the ILM allows all tools to share a single knowledge and data framework which can be built in a replicable manner from linguistic data. At the heart of our strategy is the idea that all the code for the development and training of the ILMs can be openly available, built and maintained by a community of developers following open-source practices.

However, the data needed for the training, the fine-tuned language model, and its use as the basis for the writing tools should be controlled by the Indigenous community using the governance model it finds appropriate. If we were developing the tools the traditional way, as a mix of programs, rules, and machine learning, the linguistic data would be intertwined with the code, making its separation as open-source code and governed data much more difficult.

As we discuss later, we are in the initial stages of creating writing assistants using this strategy and whether it will work is still a research question. In the prototypes of the writing assistants we are developing, we are using a mixed approach where we have used task-specific machine learning components, such as in the case of translators, completion systems, and spell checkers, and traditional procedural programming, for the dictionaries. We are doing this to accelerate the development of prototypes so they can be immediately explored and co-designed with the Indigenous communities. As we progress and the needs and requisites become clearer, we plan to replace the specifically created tools with the generic ILM model.

\begin{figure*}[t!]
    \centering
        \includegraphics[width=0.7\linewidth]{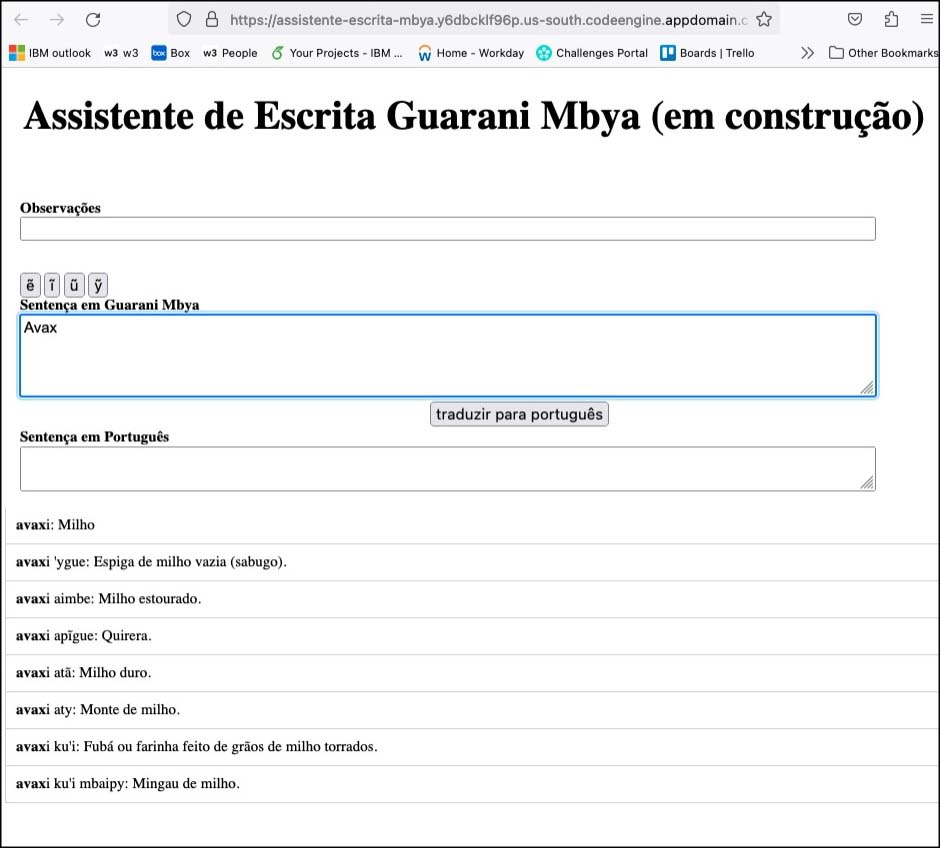}
    \caption{Prototype of the writing assistant developed for Guarani Mbya, May of 2023.}
    \label{fig:writing_assistant_guaranimbya}
\end{figure*}

\begin{figure*}[t!]
    \centering
    \begin{subfigure}{0.7\textwidth}
        \includegraphics[width=\linewidth]{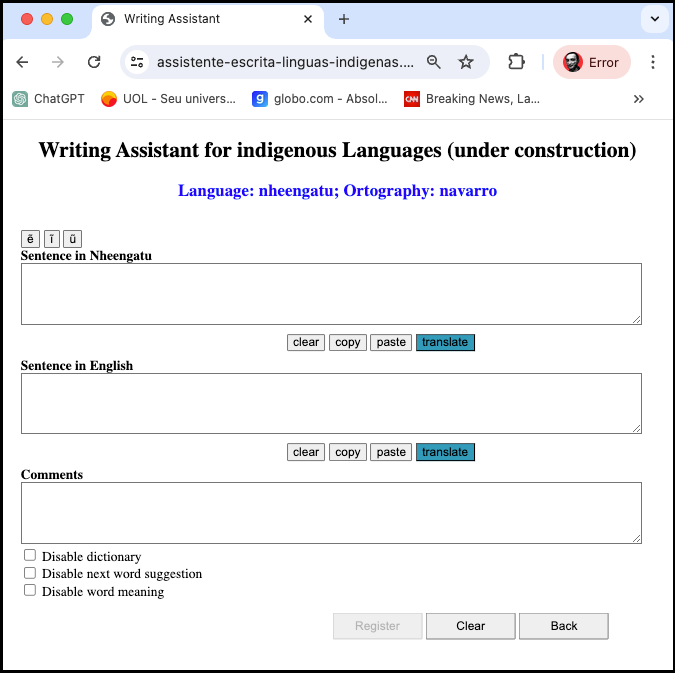} 
        \caption{main screen}
    \end{subfigure}
    \\ \vspace{2mm}
    
    \begin{subfigure}{0.48\textwidth}
        \includegraphics[width=\linewidth]{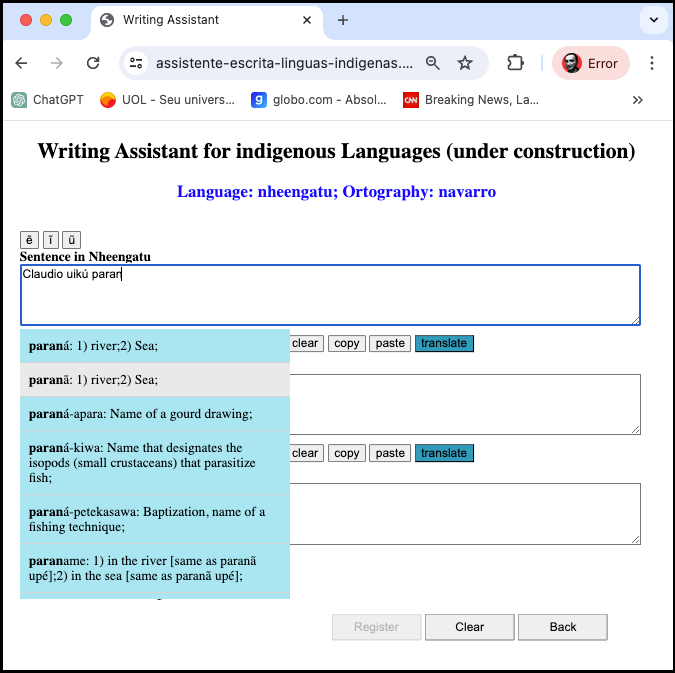}
        \caption{word completion}
    \end{subfigure} 
   \hfill
    \begin{subfigure}{0.48\textwidth}
        \includegraphics[width=\linewidth]{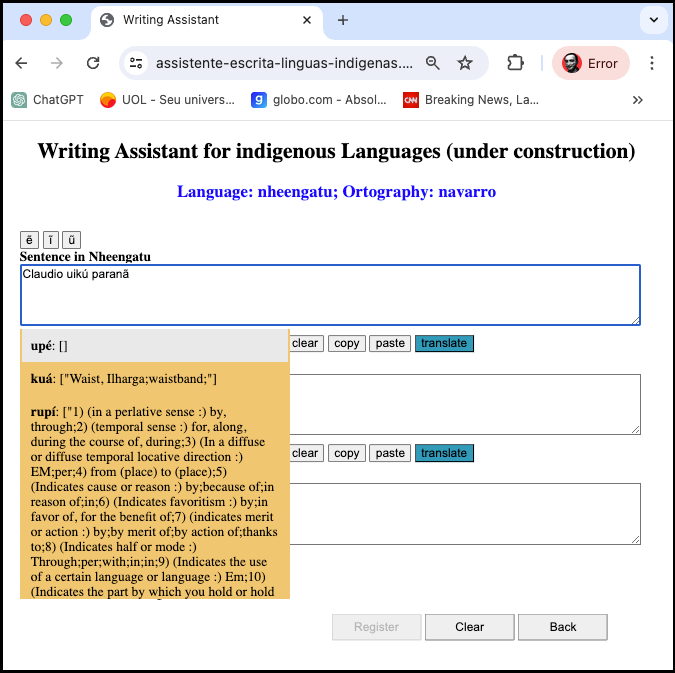}       
        \caption{next word prediction}
    \end{subfigure}
    % \hfill
    % \begin{subfigure}{0.49\textwidth}
    %     \includegraphics[width=\linewidth]{figures/writing_assistant_translation_border.png}
    %     \caption{translation}
    % \end{subfigure}

    \caption{Main screen (top) and details of the word completion (bottom left) and of the word prediction tools (bottom right) in the Nheengatu writing assistant prototyped in April of 2024.}
    \label{fig:writing_assistant_basic}
\end{figure*}

\subsection{Prototyping Writing Assistants}

The development of prototypes of writing assistants started about one month after we begun to have weekly workshops at the Tenondé-Porã high school. We employed \textit{technology probes}~\cite{hutchinson2003technology}, a variation of the idea of \textit{cultural probes}~\cite{gaver1999projected}, a.k.a. \textit{design probes}. The main idea was to insert some sort of technological artifact into the classroom which could elicit responses from the students in the context of actual writing tasks.

The first technology probe was a rudimentary version of the Guarani Mbya to Portuguese translator, accessible through a bare-bones Internet interface. It showed it had almost no use for the students, aggravated by the poor quality of the translations. But it triggered good conversations with the students about what kind of writing support would be useful for them. Based on this feedback, we started focusing on another component tools which seemed to be more useful, such as the tools to do word and sentence completion, and access to a dictionary.

\begin{figure*}[t!]
    \centering
    %\begin{subfigure}{\textwidth}
        %\includegraphics[width=1\linewidth]{figures/writing_assistant_translation_long_nhetoeng_init_border.png} 
    %\\ \vspace{2mm}
        \includegraphics[width=1\linewidth]{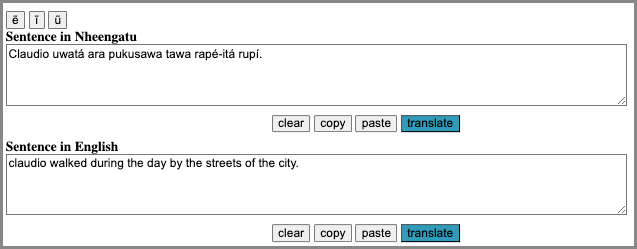}
      %  \caption{Translation: Nheengatu to English}
    %\end{subfigure} 
 
    % \begin{subfigure}{\textwidth}
    %     \includegraphics[width=0.49\linewidth]{figures/writing_assistant_translation_long_engtonhe_init_border.png} 
    % \hfill
    %     \includegraphics[width=0.49\linewidth]{figures/writing_assistant_translation_long_engtonhe_border.png}
    %     \caption{Translation: English to Nheengatu}
    % \end{subfigure} 
    \caption{Translation of a sentence from Nheengatu to English in the Nheengatu writing assistant prototyped in April of 2024.}
    \label{fig:writing_assistant_translation}
\end{figure*}

For the next iteration we created a new version of the writing assistant, incorporating early versions of those tools. The basic layout is shown in figure~\ref{fig:writing_assistant_guaranimbya}, comprising an area for writing in Guarani Mbya, a button to generate translations to Portuguese, and an area where the translation was shown. It included a word completion tool, shown in action in figure~\ref{fig:writing_assistant_guaranimbya}, suggesting possible completions for the partial word ``Avax'' as a list in the bottom of the interface, followed by their meanings in Portuguese. The sentence completion tool worked in identical ways except that was triggered by the typing of spaces and punctuation marks.

We worked with the students in a couple of workshops with this prototype.  Our anecdotal observations did not see much actual acceptance of the suggestions provided by the tools nor much use of the translator. However, the whole environment seemed to be conducive to explorations and to motivate the students to write, and  by the third workshop we started to see longer and more complex sentences being produced as expected. This in spite of the tools still having a very limited performance and many students struggling with typing using the laptops we provided (they were used to type in cell phones). The main conclusions were that the idea of a writing assistant as a tool to foster writing was valid but that better component tools were needed and, eventually, a smartphone version.

% \begin{figure*}
%     \centering
%     \begin{subfigure}{0.49\textwidth}
%         \includegraphics[width=\linewidth]{figures/writing_assistant_main_border.png} 
%         \caption{main screen}
%     \end{subfigure}
%     \hfill
%     \begin{subfigure}{0.49\textwidth}
%         \includegraphics[width=\linewidth]{figures/writing_assistant_word_completion_border.png}
%         \caption{word completion}
%     \end{subfigure} \\
 
%     \begin{subfigure}{0.49\textwidth}
%         \includegraphics[width=\linewidth]{figures/writing_assistant_next_word_border.png}       
%         \caption{next word prediction}
%     \end{subfigure}
%     \hfill
%     \begin{subfigure}{0.49\textwidth}
%         \includegraphics[width=\linewidth]{figures/writing_assistant_translation_border.png}
%         \caption{translation}
%     \end{subfigure}

%     \caption{Four screenshots of the Nheengatu writing assistant prototyped in April of 2024.}
%     \label{fig:writing_assistant_basic}
% \end{figure*}

As we moved to work with the Nheengatu language, we ported the writing assistant to Nheengatu. The main screen of this application (in its English version) is shown in figure~\ref{fig:writing_assistant_basic}, following basically the same design but with some improvements in the drop down mechanisms of the word and next-word completion. More importantly, for the Nheengatu language we developed a Portuguese to Nheengatu translator which we also made available in the interface (see figure~\ref{fig:writing_assistant_translation}).

We are in the process of introducing this version of the writing assistant to two groups of Nheengatu speakers. The first is composed of students of the State University of Campinas, mostly from the \textit{Baré} ethnic group, who have very diverse knowledge and fluency levels in Nheengatu. A preliminary workshop indicated that the usage is quite different for individuals who know and do not know how to read and write. We could also see that the use of the translators was more common, possibly because of the better quality of our Nheengatu translators (see figure~\ref{fig:writing_assistant_translation} for an example) compared to the Guarani Mbya one. We are currently organizing a writing workshop in the second half of 2024 to continue the co-design of this application.

The second group of Nheengatu speakers are Indigenous teachers and translators who have both a high command and a high need of writing in Nheengatu. This collaboration is still in its initial stages but one of our goals is to develop versions of the writing assistant which are more appropriate for writing and translation work, such as the ability to handle long texts, and investigating its  incorporation into professional or open-source word editors.

From the interaction we are having with teachers and leaders of the Indigenous communities, it is fundamental that we make the writing assistants available in smartphones. First, the majority of teenagers and young adults do not have access to traditional screen-and-keyboard computers but only to cell phones. Second, as we saw in the work with the Guarani students, most of people in this group do not have experience nor proficiency with traditional keyboards, preferring the use of thumb-based typing common in mobile phones. And finally, in the hot and humid environment of many Indigenous groups in Brazil, smartphones have been much more successful in fending off damage caused by mold and humidity.

\begin{figure*}[t!]
     \centering
     \includegraphics[width=7.8cm]{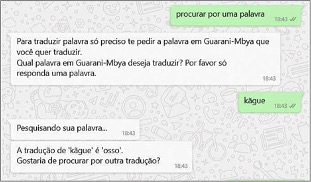}
     \hspace{2mm}
      \includegraphics[width=7.8cm ]{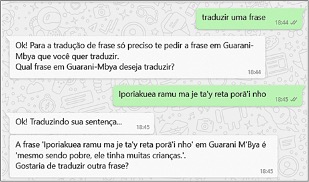} 
     \caption{Writing assistant for the WhatsApp platform: a request for the meaning of a word (left) and for a translation (right); prototype of May of 2023.}
    \label{fig:whatsapp_writing_assistant}
\end{figure*}

Considering those issues, we have been exploring, in parallel with the efforts described before, the deployment of writing assistants in more readily available platforms, such as smartphones and social media apps. Figure~\ref{fig:whatsapp_writing_assistant} shows an early prototype of the writing assistant as a chatbot in \textit{WhatsApp}, where dictionary and translation services are made available through interactions with the chatbot\footnote{This work was developed in collaboration with undergraduate computer engineering students from \textit{Insper}.}. Currently we are starting to work on porting the writing assistant to an Android app, together with a system to re-configure the keyboard we have already developed\footnote{Also, in a collaboration with students from \textit{Insper}.}.

\subsection{Developing the Component Tools}

In the development of the actual tools used in the prototypes of the writing assistants described in the previous paragraphs, we did not use a single ILM as the source of the component tools as advocated in the description of our technical strategy shown in figure~\ref{fig:technical-strategy}. The main reason was that the process of co-design and development of the prototypes required to have versions of those tools working, to some extent, in the early steps of the initialization process, as part of the new development cycle proposed before (depicted in  figure~\ref{fig:ai_cycles}.b).

We now briefly describe the development process of the tools which were actually used and discuss how we will move to a single-source ILM framework. Notice that the existence of those tools will eventually simplify the training of the generic ILM because they can be easily employed to generate prompt-like synthetic training data for the generic ILM.

The \textbf{word dictionary} used in the Guarani Mbya and Nheengatu prototypes was implemented by looking up a database of words extracted from actual dictionaries available for the  language. This database contained associated descriptions in Brazilian-Portuguese and English of each word. One shortcoming of that version was not listing valid variations of base words, such as  the use of prefixes and suffixes to indicate gender, number and verbal tenses.  Most dictionaries for major languages used in editing tools often incorporate the handling of variations by direct programming rules for valid modifications, a time- and effort-consuming process. We believe this can also be achieved by training the ILM with appropriate synthetic data, an idea yet to be verified in practice.

The \textbf{word completion} tool used in the Guarani Mbya and Nheengatu prototypes used a variation of the techniques used by the word dictionary, based on partial searches in the database of words. Partially entered words were matched against the list of valid words and the most likely words were listed in alphabetical order.

%\textbf{PAULO, PLEASE CHECK AND IMPROVE THIS}.
The \textbf{next-word completion} tool used in the Guarani Mbya and Nheengatu prototypes have been implement as a \textit{bag-of-words} machine learning model~\cite{weiss2010book}, trained with sentences extracted from the bilingual datasets created for the training of the translators, detailed before. Next-word completion is a task extremely dependent on the context and good performance is often achieved through personalization, that is, on learning the most commonly used words by an individual. In this initial version, however, we focused on a general-purpose training set, where we decomposed the original sentences in sub-sentences with up to five tokens, and used the subsequent token as the label to train an SVM-based classifier. Consequently, when the user was typing in the writing assistant, the system sent the typed words to the classifier and the predicted class corresponded to the suggested next word.

%\textbf{PAULO, PLEASE CHECK AND IMPROVE THIS}. 
We have not yet deployed a \textbf{spell checker} tool in the Guarani Mbya and Nheengatu prototypes, but one is being developed for Nheengatu using an LLM-based framework. The basic idea is to first generate a dataset with pairs of correct and incorrect sentences, where the incorrect versions are created synthetically by changing, removing, or adding letters, following common human patterns of producing typos. We first implemented this methodology with a dataset in Portuguese language, where we obtained an accuracy of about 60.8\%. We are currently in the process of applying the same methodology to  Nheengatu, using sentences extracted from the bilingual datasets created for the training of the translators, as we did with the next-word prediction tool. Besides, we are also investigating more effective ways to evaluate the results, such as applying a BLEU-like metric to compute the quality of the generate results. 

%\textbf{PAULO, PLEASE CHECK AND IMPROVE THIS}. 
The \textbf{translator} tools used in the Guarani Mbya and Nheengatu prototypes were different versions of the bilingual translators described previously in detail in section~\ref{sec:translators}. The translators were deployed as API-based services and had an average response time of 1-2~seconds, which was found adequate for the task. 

Notice that the translators we have used were  based on fine-tuning large, high-resource, high-quality translators to Indigenous languages, generating translation-only ILMs and not generic, prompt-based LLMs. Given that the accuracy of translation-specific models is today often better than of generic LLMs, it is a good question whether having a translation-specific ILM in our technical strategy would not be a better option. We will explore this issue as we start developing the generic-task ILMs proposed in the strategy. An additional issue is that translators also include a second language to translate to and from, which may introduce noise into the generic ILM. %, which could be one more reason to use of a translation-specific model.

%The spell checker technology, developed with the help of an undergrad student from MIT, is based on retraining a Large Language Model to correct words.

It is also important to highlight that the strategy of fine-tuning a single ILM to be a source of the component tools requires the development of methods and code which generate the appropriate synthetic training data, in the right format. This data can be complemented, eventually, by data logged from the writing assistants, obviously with the permission of the users and from the community, and with provisions to preserve privacy and remove personal information. We believe some of the synthetic data generators may also require language-specific knowledge which we will have to be encoded and tested by developers. 

Those data generators are part of the set of tools depicted between the data sources and the fine-tuning process in figure~\ref{fig:technical-strategy}, for which we are developing generic versions which will be eventually open-sourced, without language data. This layer is an important area of language-specific collaboration between Indigenous speakers, linguists, and developers of each language, and how to foster such collaborative environment and at the same preserve governance of the Indigenous language data is an open question.

%\subsection{Opportunities and Challenges}

%%%%%%%%%%%%%%%%%%%%%%%%%%%%%%%%%%%%%%%%%%%%%%%%%%%%%%%%%%%

\section{Can Languages Be Documented by Endangered Language Models?}

\begin{figure}[t!]
    \centering

    \includegraphics[width=7.6cm]{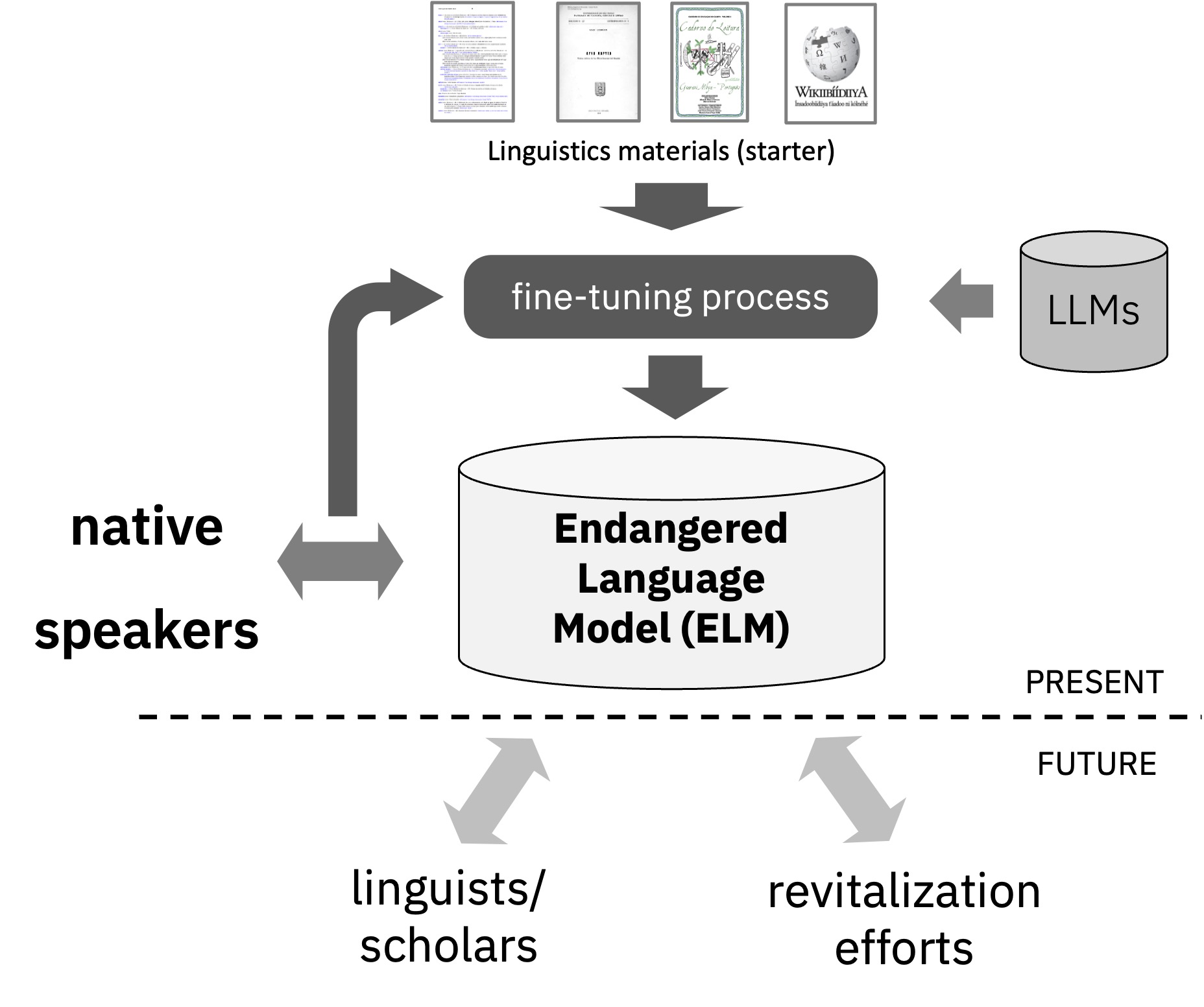}

    \caption{Diagram of an Endangered Language Model (ELM).
    }
    \label{fig:elms}
\end{figure}

When the last native speaker of a language dies, a language is lost together with the large amount of the knowledge accumulated through generations by peoples and communities. Humanity as a whole loses not only tacit knowledge about the world and the culture of communities but also different ways of thinking and structure knowledge and cognition. %The death of the last speaker of a language is a tragedy as significant as the destruction of a cultural or archaeological site or the extinction of a species.

When a language ceases to be spoken and used, what is left, in the best cases, are documents and media registering its use by previous speakers: texts, books, recordings, videos. In many cases, there are also linguistic material such as grammars, dictionaries, articles and books, and other linguistic materials with systematic, analytical descriptions of the language. In some rare cases, such media has been used to revitalize the language, that is, to enable a new generation of speakers to use the language or a good approximation of it. The most famous case of revitalization is the case of \textit{Hebrew}, which was a lost language in the beginning of the 20\textsuperscript{th} century and currently is spoken and used by millions of people~\cite{thomason2015endangered}.

The tools most commonly used today by linguists for the documentation of Indigenous languages are \textit{FLEX}\footnote{\url{https://lingtransoft.info/apps/flex-fieldworks-language-explorer}.} and \textit{ELAN}\footnote{\url{https://archive.mpi.nl/tla/elan}.}, which create and manage databases of annotated lexical and textual information. FLEX allows for a text to be automatically labelled or glossed with items from the dictionary and ELAN allows for a media archive to be transcribed and translated. Both have the possibility of synchronizing work done in parallel by more than one user. It would be ideal if these tools could be adapted computationally to interact with AI systems, if possible, and adapted to smartphones, to facilitate documentation by linguists and speakers. This improvements would be of great help on making documentation less time consuming and more accessible. 

Linguistic documentation is vital for the preservation and vitalization of Indigenous languages, and native and community speakers must be able to work actively as researchers on these databases together with non-Indigenous researchers. Moreover, efficient tools which record the time spent by users on the database could inaugurate a new economic activity for Indigenous researchers, allowing documentation projects switch from grants to hourly payments to remunerate participants.

The emergence of LLMs and interactive chatbots based on them has created a possible new framework to document an endangered language, based on the development of \textit{Endangered Language Models (ELMs)} following methodologies similar to our technical strategy described in figure~\ref{fig:technical-strategy}. Figure~\ref{fig:elms} shows the main components of the process of creating and using ELMs, including the direct interaction of the model with native speakers as a key source of training data.

We are not aware of any on-going initiative to preserve a language using this approach of creating an ELM for it but we think this is possibly a novel way to avoid the negative aspects of losing a language. ELMs may be able to provide future generations with more interactive ways to do research on a lost language, access some of the knowledge associated to it, and even revitalize it. We discuss next some of the benefits of this approach and the challenges associated with building ELMs to preserve endangered languages.

The main differences between using an ELM and the current documentation methods using medias such as books and recordings is that an ELM is more generative but less correct than traditional documentation media. Therefore, we envision here the use of ELMs not as a replacement for traditional media but as a new tool in the documentation toolbox, one that allows interactive conversation and the easy exploration of multiple cases and situations. However, it must be used with care since it may exhibit known problems of LLMs such as hallucinations, biases, and fine-tuning defects such as memorization as discussed before.

But how could such ELMs be actually built? Should the ELM be developed carefully by ML technicians guided by linguists, as a a way of cleaning the data, that is, using relevant structures which often do not appear easily in texts and recordings, such as paradigms and complex sentences? Or should the ELM be trained ``in the wild'', using the help of, for instance, \textit{reinforcement learning (RL)} methods? Not only these questions seem to drive complex technical choices but they also rely on ethical and cultural decisions and processes which are hard to foresee. 

Linguistic documentation have dealt before with issues of incorporating new technologies such as the photograph, the film, the audio and video recordings, and, more recently, the Internet and the mobile phone, and have developed new research protocols and methodologies. We envision a similar process would be necessary to establish ELMs as a new form of documentation.

Finally, it is important to consider which kinds of endangered languages are more likely to be feasibly preserved as an ELM. Going back to our analysis in section~\ref{sec:decreasing_diversity}, we do not think there is going to be enough data to train and speakers to test an ELM in the case of languages with less than 10~speakers, and even possibly in the case of 10~to 100 speakers, which we depicted as black and grey bars in the histograms of figure~\ref{fig:america-languages-histogram}. The best candidates are likely to be in the 100~to~1,000 range, about a third of the Indigenous languages in the Americas and in Brazil. For languages with more than 1,000 speakers, on-the-ground initiatives of teaching young children to speak the language seem to be a more effective way to make sure that the language does not disappear.

%%%%%%%%%%%%%%%%%%%%%%%%%%%%%%%%%%%%%%%%%%%%%%%%%%%%%%%%%%%

\section{Previous and Related Work}\label{sec:related_work}

The work described in this paper certainly did not happen in a vacuum. There have been considerable work, especially in the last 5~years, looking into how to apply AI technologies in support of endangered languages and, in particular, Indigenous languages. A detailed survey is beyond the scope of this paper.

A good survey of early work was done by Mager et al.~\shortcite{mager2018challenges} which described work, data resources, and challenges of language technologies for American Indigenous languages. Kuhn et al.~\shortcite{kuhn2020indigenous} described many different language technology initiatives of the \textit{Indigenous Languages Technology (ILT)} project at the \textit{National Research Council of Canada}, including the construction of corpora for several languages, annotation tools,  speech recognition systems, and read-along audiobooks. Neubig et al.~\shortcite{neubig2020summary} summarized a workshop on the state of use of technology for language documentation in 2019. And \citet{mager-etal-2023-neural} presented a detailed discussion on the challenges and common approaches to develop machine translation systems for Indigenous languages of America.

In particular, NLP technologies have been used in varied contexts and scenarios of endangered languages. Alavi et al.~\shortcite{alavican} discussed whether an automatic conversational system could be used to document languages;  Anastasopoulos~\shortcite{anastasopoulos2019computational} explored diverse language tools for language documentation;  Anastasopoulos et al.~\shortcite{anastasopoulos-etal-2020-endangered} discussed modern NLP issues with endangered languages; Bird~\shortcite{bird2018designing} looked into the specific issue of using mobile technologies; Cruz and Waring~\shortcite{Cruz2019DeployingTT} listed linguistic issues of using technology for endangered languages; Everson and Waring~\shortcite{everson2019online} described a platform for community-based description of Indigenous languages; Foley et al.~\shortcite{foley2018building} described the process of building speech recognition systems for language documentation; Katinskaia~\shortcite{katinskaia2017revita} presented a language learning system to support endangered languages; Maldonado et al.~\shortcite{maldonado2016ene} described a system for automatic recognition of \textit{Guarani} speech; Martín-Mor~\shortcite{martin2017technologies} explored the use of technologies for \emph{Sardinian} languages; Maxwell and Bills~\shortcite{maxwell2017endangered} discussed how digitizing print dictionaries could help to create data for endangered languages; Mirza~\shortcite{mirza2017design} explored social persuasive ubiquitous knowledge systems in the context of the \emph{Maori} language; Simha~\shortcite{simha2019} explored automatic speech recognition systems; Ubahlet~\shortcite{ubaleht2021lexeme} presented a system to manage corpora of endangered languages; Van Esch et al.~\shortcite{van2019future} explored future directions to in automatic support for language documentation; Yangarber~\shortcite{yangarber2018support} explored support for endangered and low-resource languages
via e-Learning, translation, and crowd-sourcing; and Zuckermann et al.~\shortcite{zuckerman2021lara} studied a web platform for revival and documentation based on community engagement.

Finally, the workshops on \textit{The Use of Computational Methods in the Study of Endangered Languages (ComputEL)} are a good source of real, field applications of technology to endangered languages\footnote{\url{https://computel-workshop.org/}}, as well as the workshops of the \textit{Special Interest Group on Under-resourced Languages (SIGUL)} set up by the \textit{ELRA Language Resources Association (ELRA)} and the \textit{International Speech Communication Association (ISCA)}\footnote{\url{https://www.sigul.eu/}}; and the series of workshops on \textit{NLP for Indigenous Languages of the Americas (AmericasNLP)\footnote{\url{https://turing.iimas.unam.mx/americasnlp/}}}.

\section{AI is not Enough}

This paper summarizes and discusses many different works~\cite{cavalin2023understanding,pinhanez2023balancing,cuichi2023paper,pinhanezillc24,cavalinnaacl24,dominguessigul24,vasconcelos2024} developed since 2022 under the joint project of IBM Research and the University of São Paulo to develop AI technologies to strengthen Brazilian Indigenous languages\footnote{\url{https://c4ai.inova.usp.br/research_2/\#ProIndL_B_eng}.}. This work was fundamentally the result of the discussions and collaboration we had with many Indigenous individuals and communities, particularly with the Tenondé-Porã community and the Indigenous speakers and linguists of Nheengatu. 

It would be naive of us to pretend that AI technology can save endangered Indigenous languages from decline and disappearance. First, for many Indigenous communities, the loss of their language is, regretably, the by-product of the disappearance of their own community or ethnicity. Many Indigenous communities are under threat of expulsion from their lands, of violence, and of economic exploitation. Moreover, history is adamant that languages are vitalized by their own communities, under their own practices and motives, in complex socio-political processes undertaken by their native speakers~\cite{thomason2015endangered,carrollCAREPrinciplesIndigenous2020}. 

In particular, we should be careful with the portrayal of AI tools as ``white savior'' technologies, and see the role of our work, described in this paper, as the exploration of technologies which some communities may adopt, use, and, eventually, appropriate, at their own discretion and decision.

Moreover, as discussed in the first part of the paper, the development of AI-based tools has to be done with and for the Indigenous communities, under ethical principles and relational accountability~\cite{wilsonResearchCeremonyIndigenous2008}. In the case of AI, and ML tools in particular, issues of data sovereignty, consent, and intellectual property are particularly important~\cite{harding2012conducting,walter2021indigenous,kukutai2016indigenous,kukutai2023indigenous}, although hard to communicate and to reach agreement on. Efforts to create new agreements and practices, such as the \textit{Kaitiakitanga} data license\footnote{\url{https://github.com/TeHikuMedia/Kaitiakitanga-License}} seem to be going in the right direction.

From a technical standpoint, we believe our work has been contributing in many ways to the understanding of the possibilities of AI technology in the context of vitalization of endangered Indigenous languages. We believe our proposal of an AI development cycle focused on community usage, engagement, and sovereignty, described in section~\ref{subsec:ai_cycles} is a good way to address some of the key ethical issues discussed in section~\ref{subsec:ai_research_indigenous}. In section~\ref{sec:translators}, we have shown that fine-tuning machine translators with ultra-low amounts of data has to be done with extreme care, due both to memorization effects and the dangers of contamination. We also provided evidence that data quantity matters and, more importantly, that data quality is essential.

Our work with writing assistants described in section~\ref{sec:writing-assistant}, although in its early stages, is developing methodologies and practices for co-designing writing-support tools for Indigenous communities and exploring the feasibility of a technical strategy we proposed based on Indigenous Language Models (ILMs). 
Finally, we are proposing here a new form of language documentation based on Endangered Language Models (ELMs) where some of the technologies developed for translators and writing assistants may be applicable. 

Overall, we have been developing a set of technologies and methods which explore how to harness the recent advances in AI to support the vitalization of endangered Indigenous languages.
The use of the term ``harness'' here and in the title of this paper is intentional, because the two meanings of the term represent well our main goals. ``To harness'' can be understood as ``to control'', which is a good way to view our efforts to use fine-tuning mechanisms to harness high-resource MTs to produce translations of Indigenous languages, or the methodology we are using to develop spell checkers. But ``to harness'' can also mean ``to utilize'' the power of AI, and particularly of modern NLP technologies such as LLMs, to create tools that would otherwise be impossible or very difficult to build.

The research described here should be regarded as the initial results of our project. We are still in the beginning of our engagement with the Nheengatu writers and translators and we expect many new design and development ideas to be created as we work together and see those and new tools being developed. Equally, the development and use of writing assistants are in the infancy, and we hope that the coming workshops with Nheengatu-speaking students will take the design and deployment of those tools in new directions. We are also exploring, with other Indigenous communities and related organizations, how such tools could be used in other contexts, such as law, health, and education, and in the support of economic activities.

From a technical view, we are continuing the exploration of techniques to build better translators, exploring different methods of fine-tuning, and looking into how to produce synthetic data. We also want to work with other research groups to allow them to use our translator-development tools to create translators for other languages, to ensure that the technology is replicable. In the case of writing assistants, our main goal is to finish the development of a first, complete, usable prototype which can be used and tested by a large and diverse community of users, delivered from multiple platforms. The next step is to move towards a single-source ILM able to provide the functionality of all the component tools (except, possibly, translation, as discussed before). The following step would be the creation of a set of basic training data generators, closing the ILM production cycle, and then work with multiple communities to use and enhance the technological framework.

During the first day of our first workshop at the Tenondé-Porã high school, one of the students asked us when we would start teaching them computer science. We are aware that this project will not be successful in producing real impact if the technologies developed are not transferred and appropriated by the Indigenous communities, as part of the ``community sovereingty'' long-term process of the proposed AI development cycle (see figure~\ref{fig:ai_cycles}.b). A parallel effort is needed to cultivate Indigenous linguists, computer scientists, programmers, and UX designers to ensure the continuity of successful tools and technologies.

Many Indigenous leaders we have talked to are cognizant of the importance of empowering   their communities with digital tool-making technologies. The adoption, adaptation, and appropriation of new technologies by non-Indigenous people have been the history and the reality for most of these communities for the last 500~years. In many ways, our project will be really successful when our involvement is no longer necessary, and Indigenous communities are fully empowered to independently develop, utilize, manage, and govern these language and AI technologies.
% We will succeed in this project when we are not needed anymore.

\section*{Acknowledgements}

%\textbf{EVERYONE PLEASE CHECK AND INCLUDE NAMES IN THIS SECTION}

Most and foremost, we acknowledge the vast amounts of knowledge and wisdom we have generously received from the people we worked with in the Indigenous communities, from their leaderships, and from people of NGOs who work with them: the teachers and students of the Gwyra Pepo Guarani School, in the Tenondé Porã community, in particular Jordi Karai Mirim and Joana Cabral de Oliveira, and from Talita Lazarin Dal Bo and Ana Luísa Brites Blaser from the \textit{``Programa de Educação do Comitê Interaldeias''}; from the Nheengatu community, Arlindo Baré, Claudia Baré, Elizângela Baré, Cauã Borari, Édson Baré Warixi, Melvino Baré, and Marivelton Barroso Baré; and from the Terena community, Dario Terena. We also acknowledge the work of undergraduates who were involved at different stages of this work: Caio Medeiros, Marina Geiger, and Sofia Cabral from the University of São Paulo, and Carolina Souza from MIT.

We also want to thank the people who have been worked with us in this project but whose work is not directly reflected in the works described here: professors Sarajane Peres, Antônio Cândido, and Marcelo Finger from the University of São Paulo, and Karin Vivanco from the State University of Campinas; and the student Gustavo Evangelista.

We also thank many people who have allowed us to access results, data, and publications related to Indigenous languages: prof. Robert Dooley, prof. Marcel D'Avila, and prof. Eduardo Navarro.

Finally we acknowledge the leadership of the institutions who have supported this work: Bruno Flach and Stacy Hobson from IBM Research, professor Fábio Cozman of the Center for Artificial Intelligence (C4AI) of the University of São Paulo, and Rosabelli Coelho-Keyssar of the MIT Brazil Program. 

This work has been partially funded by the Center for Artificial
Intelligence (C4AI-USP), with support by the FAPESP (São Paulo Research Foundation, grant \#2019\/07665-4) and by the IBM Corporation.

%\bibliographystyle{acl}
%\bibliography{full_indigenous}

\end{document}